\documentclass[12pt,twoside,a4paper]{article}
\usepackage{amssymb}
\usepackage{amsthm}
\usepackage{amsmath}
\usepackage{graphicx}
\usepackage{multicol}
\usepackage[nofoot,top=30mm,left=3.00cm,right=3.00cm,bottom=10mm]{geometry}
\usepackage{epsfig}
\usepackage{cite}
\usepackage{fancyhdr}
\begin{document}
\pagestyle{fancy}
\rhead[Digital Image Processing - A Review]{\thepage}
\chead{}
\lfoot{}\cfoot{}\rfoot{}
\thispagestyle{empty}
\lhead[\thepage]{K.P.N.Murthy, M. Janani and B. Shenbaga Priya}
\rhead[Markov Chain Monte Carlo Image Restoration]{\thepage}
\begin{center}
{\bf\Large Bayesian Restoration of Digital Images Employing}
\vskip 3mm
{\bf\Large Markov Chain Monte Carlo - a Review}\\[5mm]
K. P. N. Murthy${}^{\dagger}$,
M. Janani${}^{\star}$ and B. Shenbaga Priya${}^{\star}$\\[5mm]
${}^{\dagger}$ School of Physics, University of Hyderabad,\\
Central University P.O.,
Hyderabad 500 046,\\
Andhra Pradesh, INDIA.\\
${}^{\star}$ WIPRO Technologies,
475 A, Old Mahapalipuram Road,\\
Sholinganallur, Chennai 600 032,\\
 Tamilnadu, 
INDIA
\end{center}
\vskip 5mm
\begin{abstract}
A review of Bayesian restoration of digital images based on 
Monte Carlo techniques is 
presented. The topics covered include  Likelihood, Prior and 
Posterior distributions, Poisson, Binary symmetric channel and Gaussian 
channel models of Likelihood distribution, 
Ising and Potts spin models of Prior distribution, 
restoration of an image through Posterior maximization, statistical 
estimation of a true image from Posterior ensembles, Markov Chain Monte Carlo
methods and cluster algorithms.   
\end{abstract}
\vskip 5mm

\section*{Introduction}
Noise, in a digital image, is an inevitable  nuisance; 
often it defeats the
purpose, the image was taken for. 
Noise mars the scene it intends to depict and affects 
the artist in the person
looking at the image; it limits the diagnosis a doctor 
makes from a medical image; it
interferes with the inferences a physicist draws from 
the images acquired by his 
sophisticated spectrometers and microscopes; it 
restricts  useful information that can 
be extracted  from the images transmitted by 
the satellites on earth's resources and 
weather; {\it etc.} Hence it is important 
that noise be eliminated and true image  restored.

Noise, in the first place,  enters a digital 
image while being acquired  due to, for example  
faulty apparatus and  poor statistics; 
it  enters while storage  due to aging,  defects in  storage devices
{\it etc.}; it  enters 
during transmission through noisy channels.  We shall not be concerned,
in this review, 
with specific details of   mechanisms of noise entry and image degradation.  
Instead, we  confine our attention mostly to  de-noising,  also called 
restoration, of digital images. 

Conventional digital image processing techniques rely  on the 
availability of qualitative and quantitative information about the 
specific degradation process that leads to noise in an   
image. These algorithms  work efficiently on  
applications they are intended for; 
perhaps they would  also work reasonably well 
on closely related applications. But they
are unreliable for applications falling 
outside\footnote{Often the mechanism of image degradation is 
 not known reasonably well. In several situations, even if the 
mechanism is known, it is complex and  not amenable to 
easy modeling. It is in these contexts that general purpose 
image restoration algorithms become useful.} their domain.
There are not many 
image restoration algorithms that can handle a wide variety of images. 
Linear filters like low pass filters, high pass filters  {\it etc.},  and 
non-linear filters like median filters, constrained least mean square filters {\it etc.},
are a few  familiar examples belonging to the category of general purpose algorithms.

An important  aim of image restoration research  
is to devise robust algorithms based on 
simple and easily implementable ideas and which can 
cater to a wide spectrum of applications. 
It is precisely in this context,
a study of the statistical mechanics of images 
and image processing should
prove useful. The reasons are many. 
\begin{enumerate}
\item[-]
Statistical mechanics aims to describe macroscopic properties  of a material
from those of its microscopic constituents and their interactions;
an image is a macroscopic object and its microscopic constituents are
the gray levels in the pixels of the image frame.
\item[-]
Statistical mechanics is based on a few general and simple principles and 
can be easily adapted to image processing studies.
\item[-]
Bayesian statistics is closely related to 
statistical mechanics as well as to the foundations of 
stochastic image processing in particular and 
information processing in general.
\item[-]
Several statistical mechanical models  
have  a lot in common with image processing models. 
\end{enumerate}

But why should we talk of these issues now ?
Notice,  images have entered our personal 
and professional life in a big way
in the recent times - personal computers, 
INTERNET, digital cameras, mobile 
phones with image acquisition and 
transmission capabilities, new and sophisticated 
medical and industrial imaging devices, 
spectrometers, microscopes, remote sensing
devises {\it etc.}.

Let me  briefly touch upon a few key  
issues of  image processing to drive home the
connection between statistical mechanics 
and image processing.
Mathematically, an  image is a matrix of 
non-negative numbers, representing the gray levels or 
intensities in the pixels of an image plane. A pixel is  a tiny square 
on the image plane and represents an element of the picture. Pixels 
coarse-grain the image plane. The gray level in 
a pixel defines the state of the pixel.
Pixels  are analogous 
to the vertices of a lattice that coarse-grain space 
in a statistical mechanical two dimensional  model. The gray levels
are analogous to  the states of spins or of 
atoms sitting   on the lattice sites.
Spatial correlations in an image can be modeled by defining suitable 
interaction between states of two pixels, exactly 
like the spin-spin interaction 
on a lattice model.
In fact one can define an energy for  an image by  summing over these  
 interactions. 
An exponential Bayesian {\it Prior}  is analogous to  Gibbs  distribution.
It is indeed this correspondence 
which renders an image, a Markov random field and {\it vice versa}. 
We can define a Kullback-Leibler entropy 
distance for constructing a {\it Likelihood} probability distribution 
of images. Temperature, in the context of image 
processing,  represents  a parameter 
that determines the smoothness of an image.  
The energy-entropy competition, 
responsible for different phases of matter, 
is like  the {\it Prior} - {\it Likelihood} 
competition that determines the smoothness 
and feature-quality of an image in Bayesian Posterior maximization
algorithms. 
Simulated annealing that helps equilibration in statistical 
mechanical models has been employed in image processing
by way of starting the restoration process at high temperature
and gradually lowering the temperature to improve
noise reduction capabilities of the chosen algorithm. 
Mean-field approximations of statistical mechanics 
have been employed for hyper-parameters estimation 
in image processing. In fact several models and methods
in statistical mechanics have their counter parts in image processing.
These include  Ising and Potts spin models, random spin models,
Bethe approximation,  Replica methods, renormalization techniques,
methods based on bond percolation clusters and mean field approximations. 
We can add further to the list of common elements of image processing  
and statistical mechanics.  
Suffice is to say that these two seemingly disparate disciplines have 
a lot to learn from each other. 
It is the purpose of this review to elaborate on these 
analogies and describe several image restoration algorithms inspired 
by statistical mechanics. 

Image processing is a vast   field of research 
with its own idioms and sophistication. 
It would indeed be impossible  to address all 
the issues of  this subject in a single 
review; nor are we  competent to undertake 
such a task. Hence  in this 
review we shall restrict attention to a very 
limited field of image processing namely  
the Bayesian de-noising of images employing 
Markov chain Monte Carlo methods. We have made a conscious 
effort to keep 
the level of discussions  as simple as possible. 

Stochastic image processing based on Bayesian methodology was pioneered by 
Derin, Eliot, Cristi and Gemen \cite{DECG}, Gemen and Gemen \cite{SGDG} 
and  Besag \cite{JEB_1986}. For an early review
see Winkler \cite{GW}; for a more recent review on the 
statistical mechanical approach to image processing
see Tanaka \cite{KT}. Application of analytical methods of statistical mechanics
to image restoration process can be found  in \cite{JMPADB}. 
The spin glass theory of statistical mechanics has been
applied to information processing \cite{HN}. Infinite range
Ising spin models have been applied to image restoration employing 
replica methods \cite{HNKMYW}. A brief account of 
the Bayesian restoration of digital images employing 
Markov chain Monte Carlo methods was presented in \cite{KPN_Madurai}.
The present  review, is essentially based on \cite{KPN_Madurai} with 
more details and examples. The review  is organized as follows.

We start with a mathematical representation and a 
stochastic description of an image. We discuss 
a Poisson model for image degradation to construct a 
{\it Likelihood} distribution. We show how Kullback-Leibler 
distance emerges naturally  in the context of 
Poisson distributions. 
Then we show how to quantify the smoothness
of an image through an energy function. This is followed by 
a discussion of Ising and Potts spin Hamiltonians often 
employed in modeling of magnetic transitions.
We define a Bayesian 
{\it Prior} as a Gibbs distribution. Such a description renders an 
image a Markov random field. Bayes' theorem combines a subjective
{\it Prior} with a {\it Likelihood} model incorporating the data on
the corrupt image 
and gives a {\it Posterior} - a conditional distribution from which we can
infer the true image. We show how the {\it Likelihood} and the 
{\it Prior} compete to increase the {\it Posterior}, the competition 
being tuned by  temperature. The {\it Prior}
tries to smoothen out the entire image with the {\it Likelihood} competing
to preserve all the features, genuine as well as noisy, of 
the given image. 
For a properly tuned temperature the {\it Likelihood} loses 
on noise but manages to win in  preserving image features; similarly
the {\it Prior} loses on smoothening 
out the genuine features  of the 
image but wins on smoothening out the noise inhomogeneities. 
The net result of the 
competition is that we get a smooth image with all genuine
inhomogeneities  in tact.
The image that maximizes the {\it Posterior} for a given temperature
provides a good estimate of the true image. We describe a simple 
Monte Carlo algorithm to search for the Posterior maximum 
and present a few results of processing 
of noisy toy and benchmark images. 

Then we take up a detailed discussion 
of Markov Chain Monte Carlo methods  based
on Metropolis and Gibbs sampler; we  show how to estimate the true image
through Maximum $\grave{ {\rm a}}$ Posteriori (MAP), Maximum Posterior
Marginal (MPM) and Threshold Posterior Mean (TPM) statistics 
that can be extracted from an 
equilibrium ensemble populating the asymptotic 
segment of a Markov chain of images. We present results 
of processing of toy and benchmark images employing the algorithms 
described in the review. Finally we take up the 
issue of sampling from a {\it Prior} distribution employing 
bond percolation clusters. The cluster algorithms can be adapted to
image processing by sampling cluster gray level independently from the 
{\it Likelihood} distribution. Cluster algorithms ensure faster convergence
to {\it Posterior} ensemble from which the true image can be inferred. 
We employ a single  cluster growth algorithm to process toy and benchmark images
and discuss results of the exercise. 
We conclude with a brief summary and a discussion of 
possible areas of research in this exciting field of image processing,
from a statistical mechanics point of view. 

\section*{Mathematical Description of an Image}  
\rhead[Mathematical and Statistical Description]{\thepage}
\rhead[Markov Chain Monte Carlo Image Restoration]{\thepage}
Consider a region of a plane discretized into 
tiny squares called picture elements or  
pixels for short. Let ${\cal S}$ be a finite set 
of pixels. For convenience 
of notation we identify the spatial location of 
a pixel in the image plane by 
a single index $i$.
Also we say that  index $i$ denotes  a pixel. 
Mathematically,  a collection 
$\Theta=\{ \theta_i\ :\ i\in {\cal S}\}$ 
of non-negative numbers is called an image. 
Let $N\ <\ \infty$ denote  the cardinality of 
the set ${\cal S}$. In other words $N$ is the number of pixels 
in the image plane. 
$\theta_i$ denotes the intensity or gray level in the 
pixel $i\in{\cal S}$. The state of a pixel is defined by its gray level.
Let us denote by $\widehat{\Theta} = \{ \widehat{\theta}_i\ :\ i\in{\cal S}\}$, 
the true image. $\hat{\Theta}$ is not known to us.
Instead we have a {\it noisy} image $X=\{ x_i : i\in{\cal S}\}$. 
The aim is to restore the true image $\hat{\Theta}$ from the given noisy
image $X$.

\section*{Stochastic Description  of an Image}
In a stochastic description,  we consider an image $\Theta$ as a collection of 
random variables $\{ \theta_i\ :\ i\in{\cal S}\}$\footnote{ 
Equivalently, we can consider an image as
a particular  realization of the set of random variables. 
For convenience of notation we use the same symbol 
$\theta_i$ to denote the random variable as well
as a realization of the random variable. 
The distinction would be clear from the context.}.
Consider for example, $N$ pixels painted with 
$Q$ gray levels, $\{ 0,\ 1,\ \cdots ,\ Q-1\}$. 
The gray level
Label $0$ denotes black and the 
gray level label $Q-1$ denotes white.  
Each pixel can take any one of
the $Q$ gray levels. Collect all the  possible images
you can paint,  in a set, denoted by the symbol $\Omega$, 
called  the state space. $\Omega$ is analogous 
to the coarse grained phase space 
of a classical statistical mechanical system. 
Let $\widehat{\Omega}$ denote the total 
number of images in the state space. It is clear that
 $\widehat{\Omega}=Q^N$.   
Both  $\widehat{\Theta}$ and $X$ belong to $\Omega$.
If all the images belonging to
$\Omega$ are equally probable then we 
can resort to a uniform  ensemble description.

Figure (1 : Left) depicts a $10\times 10$ image painted 
randomly with $5$ gray levels labeled by integers from
$0$ to $4$, with the label $0$ denoting black and the label $4$ denoting white. 
The $10\times 10$ matrix of non-negative numbers 
(the gray level labels) that provides a 
mathematical representation of the image is also shown in Fig.~(1~:~Right).
\begin{figure}[ptb]
\label{random_image}
\includegraphics[height=2.4in,width=2.4in]{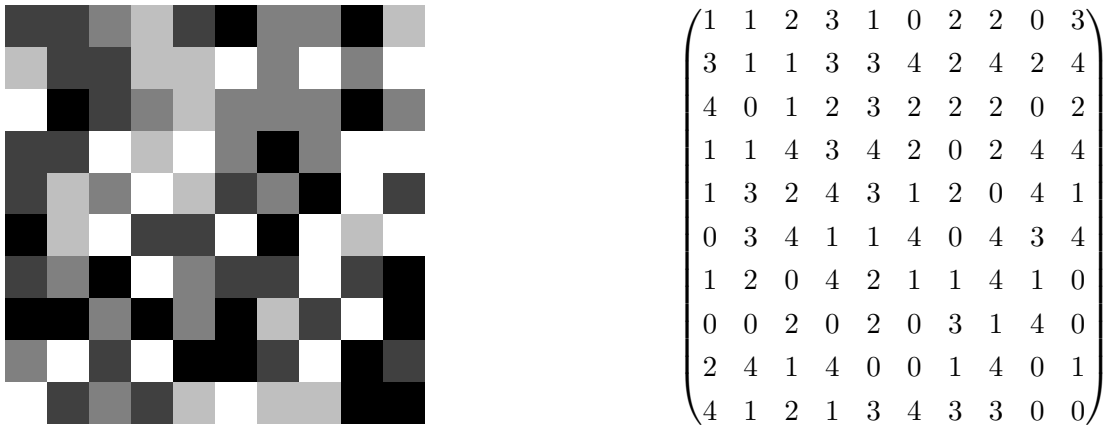}
\vglue -65mm
{\small 
\begin{eqnarray}\nonumber
\ \ \ \ \ \ \ \ \ \ \ \ \ \ \ \ \ \ \ \ \ \ \ \ \ \ \ \ \ \ \ \ \ \
\ \ \ \ \ \ \ \ \ \ \ \ \ \ \ \ \ \ \ \ \ \ \ \ \ \ \ \ \ \ \ \ \ \ 
\begin{pmatrix}
         1 & 1 & 2 & 3 & 1 & 0 & 2  & 2 & 0 & 3\\[1mm]
         3 & 1 & 1 & 3 & 3 & 4  & 2 & 4 & 2 & 4\\[1mm]
         4 & 0 & 1 & 2 & 3 & 2  & 2 & 2 & 0 & 2\\[1mm]
         1 & 1 & 4 & 3 & 4  & 2 & 0 & 2 & 4 & 4\\[1mm]
         1 & 3 & 2 & 4 & 3  & 1 & 2 & 0 & 4 & 1\\[1mm]
         0 & 3 & 4 & 1 & 1  & 4 & 0 & 4 & 3 & 4\\[1mm]
         1 & 2 & 0 & 4 & 2  & 1 & 1 & 4 & 1 & 0\\[1mm]
         0 & 0 & 2 & 0 & 2 &  0 & 3 & 1 & 4 & 0\\[1mm]
         2 & 4 & 1 & 4 & 0 &  0 & 1 & 4 & 0 & 1\\[1mm]
         4 & 1 & 2 & 1 & 3 &  4 & 3 & 3 & 0 & 0
\end{pmatrix}
\end{eqnarray}
\vskip 5mm
\caption{\small {\bf (Left) A $10\times 10$ random image painted with five gray levels with 
$0$ representing black and $4$ representing the white.
(Right) The  matrix of non-negative numbers (gray levels) representing the image.}
}}
\end{figure}
In image processing terminology,
$\Theta$ is called a random field\footnote{We shall see later 
that in a stochastic 
description, an  image 
can be modeled as a Markov random field.}.
 Thus the noisy image $X$ on hand is a 
random field. 
The noisy image $X=\{ x_i\ :\ i \in {\cal S}\}$ has 
\begin{enumerate}
\item[-] 
a systematic component representing the features of  
$\widehat{\Theta}$, that has survived  image degradation 
\item[]
and 
\item[-]
a stochastic component that accounts for the noise that enters $\widehat{\Theta}$
during degradation \footnote{the degradation of an image can occur during acquisition,
storage or transmission through noisy channels.}.
\end{enumerate}
A stochastic model for $X$ is constructed by defining a conditional probability of 
$X\vert\widehat{\Theta}$ ($X $ given $\widehat{\Theta}$). 
This conditional probability is denoted by
${\cal L}(X\vert\widehat{\Theta})$ and is called the {\it Likelihood} probability distribution,
or simply the {\it Likelihood}. ${\cal L}(X\vert\widehat{\Theta})$ can be suitably modeled 
to describe the stochastic degradation process that produced $X$ from $\widehat{\Theta}$. 

Let us consider a simple and general model of 
image degradation based on 
Poisson statistics, which will render transparent the  idea and use 
of likelihood distribution. 
\rhead[Poisson Likelihood and Kullback-Leibler Entropy]{\thepage}
\rhead[Markov Chain Monte Carlo Image Restoration]{\thepage}
\subsection*{Poisson Likelihood Distribution}
We consider the random field $X$ as constituting a collection of independent 
random variable $\{ x_i\ :\ i\in {\cal S}\}$. We take each 
$x_i$ to be  a Poisson random variable. The motivation is clear. Image acquisition
is simply a counting process\footnote{We count the  photons incident on a photographic  or X-ray film;
we count the electrons while recording   
diffraction pattern; {\it etc}.}.
Poisson process 
is a simple and a natural  model for describing counting.   
There is only one parameter in a Poisson distribution, namely the mean. 
We take $\langle x_i\rangle =\widehat{\theta}_i$. The 
{\it Likelihood} of $x_i\vert\widehat{\theta}_i$ is given by,
\begin{eqnarray}
{\cal L}(x_i\vert\widehat{\theta}_i) &=& \frac{1}{x_i\  !}
\bigg( \widehat{\theta}_i \bigg)^{x_i}\exp(-\widehat{\theta}_i)
\end{eqnarray}
We assume $\{ x_i\ :\ i\in{\cal S}\}$ to be  independent random variables. 
The joint distribution is then the  product of the distributions of the individual 
random variables. Hence we can 
write the {\it Likelihood} distribution of $X\vert\widehat{\Theta}$ as,
\begin{eqnarray}
{\cal L}(X\vert\widehat{\Theta}) &=&\prod_{i\in{\cal S}}
                                     {\cal L}(x_i\vert\widehat{\theta}_i)\nonumber\\
                                  & & \nonumber\\
                                 &=&\prod_{i\in{\cal S}}
                                    \frac{ \bigg( \widehat{\theta}_i \bigg)^{x_i}
                                    e^{-\widehat{\theta}_i}}
                                    {x_i\  !}
\end{eqnarray}
The above Poisson {\it Likelihood} can be conveniently expressed as,
\begin{eqnarray}
{\cal L}(X\vert\widehat{\Theta}) &=&
   \exp\bigg[ -F(\widehat{\Theta},X)\bigg]\ 
\prod_{i\in{\cal S}}\ \frac{1}{ x_i\  !},
\end{eqnarray}
where,
\begin{eqnarray}\label{function_F}
F(\widehat{\Theta},X)=\sum_{i\in {\cal S}}\bigg[ \widehat{\theta}_i-x_i\log(\widehat{\theta}_i)\bigg]
\end{eqnarray}
The advantage of Poisson and related likelihood distributions is that there is no hyper-parameter
that needs to be optimized\footnote{The Gaussian channel {\it Likelihood}  has two hyper-parameters.
The binary symmetric channel {\it Likelihood} has one hyper-parameter. We shall discuss
these types of degradation processes later.}.

\subsection*{Kullback-Leibler Entropy Distance}
It is clear from the discussion on  
Poisson {\it Likelihood} that  the function $F(\Theta,X)$
serves to quantify how far away is an  image $\Theta$ from another 
image $X$. Let us investigate as to what extent does the function
$F(\Theta,X)$ meet the requirements of a distance metric.  

$F(\Theta,X)\ne 0$ when $\Theta=X$ and it is not symmetric in its arguments.
Let us write $F(\Theta,X)$ as,
\begin{eqnarray}
F(\Theta,X) = \sum_{i\in{\cal S}} f(\theta_i,x_i),
\end{eqnarray} 
and study the behaviour of $f(\theta_i,x_i)$ 
as a function of $\theta_i$
with a known value of $x_i$. We have,
\begin{eqnarray}
f(\theta_i;x_i)&=&\theta_i -x_i\log(\theta_i),\nonumber\\
               & & \nonumber\\
\frac{ d^{} f}{d\theta_i}&=&1-\frac{x_i}{\theta_i}.
\end{eqnarray}
It is clear from the above that,
\begin{eqnarray}
\frac{ d^{} f}{d\theta_i} & = & 0 
\end{eqnarray}
when $\theta_i=x_i$. In other words at $\theta_i = x_i$, the 
function $f(\theta_i; x_i)$ is an extremum. To determine whether it is 
a minimum or a maximum we consider the second derivative of 
$f(\theta_i ; x_i)$ at $\theta_i=x_i$. We have,
\begin{eqnarray}
\frac{ d^2 f}{d\theta_i ^2}
\bigg\vert_{\theta_i=x_i}&=&\frac{1}{x_i}\ \  >\ \  0,
\end{eqnarray}
implying that $f(\theta_i;x_i)$ is  minimum at  $\theta_i=x_i$.
Let $f_m=x_i-x_i\log(x_i)$ denote the minimum value of 
$f(\theta_i;x_i)$.
We define a new function,
\begin{eqnarray}
g(\theta_i,x_i)&=& f(\theta_i,x_i)-f_m\nonumber\\
               & & \nonumber\\
               &=& (\theta_i-x_i)+(\theta_i-x_i)\log(\theta_i/x_i) .
\end{eqnarray}
The function $g(\theta_i,x_i)$ is not symmetric in its arguments.
Hence we define,
\begin{eqnarray}
F_i (\theta_i,x_i)&=&g(\theta_i,x_i)+g(\theta_i\to x_i,x_i\to \theta_i)\nonumber\\
                  & & \nonumber\\
               &=&(\theta_i-x_i)\log(\theta_i/x_i)
\end{eqnarray}
We can now define a  new distance between $\Theta$ and $X$ as
\begin{eqnarray}\label{Kullback_Leibler_distance}
F(\Theta,X)=\sum_{i\in{\cal S}}(\theta_i-x_i)\log(\theta_i/x_i)
\end{eqnarray}
We see that that the new distance function 
 $F(\Theta,X)=0$ 
when $\Theta=X$ and is positive definite when 
$\Theta\ne X$; also $F$ is symmetric in its arguments.
We shall take $F(\Theta,X)$
as a measure of the distance between the images $\Theta$ and $X$
\footnote{strictly  
$F(\Theta,X)$, given by Eq. (\ref{Kullback_Leibler_distance}) is not 
a distance metric  because it does not obey  triangular inequality.}.
$F(\Theta,X)$ is  the same as the 
symmetric Kullback-Leibler entropy \cite{Kullback}. This is also called by various names like
Kullback-Leibler distance, Kullback-Leibler divergence, 
mutual information, relative entropy {\it etc.}, defined 
for quantifying the separation between two probability distributions. 
We refer to $F(\Theta,X)$ given by Eq. (\ref{Kullback_Leibler_distance}) 
as Kullback-Leibler distance.

\subsection*{Hamming Distance}
\rhead[Gaussian and Binary Channels]{\thepage}
\rhead[Markov Chain Monte Carlo Image Restoration]{\thepage}
Another useful function that quantifies the separation  between two images $\Theta$ and $X$,
defined on a common image plane,
is the Hamming distance, defined as,
\begin{eqnarray}
F(\Theta, X)=\sum_{i\in{\cal S}} {\cal I}(\theta_i\ne x_i),
\end{eqnarray}
where the indicator function, 
\begin{eqnarray}
{\cal I}( \eta )=\begin{cases}
                           1\ {\rm if\ the\  statement}\ \  \eta \ {\rm is \ true},\cr
                            \ \ \ \cr                                        
                           0\ {\rm if\ the\  statement}\ \  \eta\ {\rm is\ not\ true}.
                          \end{cases} 
\end{eqnarray}  
Hamming distance simply counts the number of pixels in $\Theta$ having 
gray levels different from those in the corresponding pixels in $X$. 
The Hamming distance $F(\Theta,X)$ 
defined above is symmetric in its arguments; also
$F(\Theta,X)=0$, when $\Theta=X$.
We shall find use for  Hamming distance while 
processing binary images or even for processing images 
which have a few number of gray levels. 
However if the number of gray levels in an image is large, 
then the Kullback-Leibler distance 
should prove more useful. 

\subsection*{Gaussian Channel Likelihood Distribution}
A popular model for image degradation is based on assuming $x_i$ to be 
Gaussian random variable with 
\begin{eqnarray}
\langle x_i\rangle & = & \alpha\ \widehat{\theta}_i,\nonumber\\
\ & & \nonumber\\
\langle x^2_i\rangle-\langle x_i\rangle^2 & = & \sigma^2.
\end{eqnarray}
In the above $\alpha$ and $\sigma$ are called the 
hyper-parameters of the Gaussian channel model.
The {\it Likelihood} of $x_i\vert\widehat{\theta}_i$ is given by,
\begin{eqnarray}
{\cal L}(x_i\vert\widehat{\theta}_i)=\frac{1}{\sigma\sqrt{2\pi}}
\  \exp\bigg[ -\frac{1}{2\sigma^2}\ \Bigg(x_i-\alpha\theta_i\Bigg)^2\ \bigg]
\end{eqnarray}
We assume $\{ x_i\ :\ i\in{\cal S}\}$ to be independent random variables.
The {\it Likelihood} in the Gaussian channel model is given by,
\begin{eqnarray}
{\cal L}(X\vert\widehat{\Theta})&=&\frac{1}{\Big(\sigma\sqrt{2\pi}\Big)^N}
\exp\bigg[ -\frac{1}{2\sigma^2}\sum_{i\in{\cal S}}\Big(x_i-\alpha\theta_i\Big)^2\bigg],
\end{eqnarray} 
where $N$ is the total number of pixels in the image plane. 

\subsection*{Binary Symmetric Channel Likelihood Distribution}
Consider a binary image $\widehat{\Theta}$ which degrades to $X$ by the following 
algorithm. Take a pixel $i\in{\cal S}$ in the 
image plane of $\widehat{\Theta}$.  Call a random number $\xi$ uniformly distributed in the
unit interval. If $\xi\le\tilde{p}$ set $x_i$ to the other gray level. 
Otherwise set $x_i=\widehat{\theta}_i$. This process is described by the {\it Likelihood}
distribution given by,
\begin{eqnarray}
{\cal L}(x_i\vert\widehat{\theta}_i)&=&\tilde{p}\ {\cal I}(x_i\ne\widehat{\theta}_i)+
                    (1-\tilde{p})\ {\cal I}(x_i=\widehat{\theta}_i)
\end{eqnarray}
It is convenient to express the above {\it Likelihood} as,
\begin{eqnarray}
{\cal L}(x_i\vert\widehat{\theta}_i)&=&\frac{\exp\Big[-\beta_L {\cal I}(x_i\ne\widehat{\theta}_i)\Big]}
                                            {1+\exp(-\beta_L)}
\end{eqnarray}
The hyper-parameter $\beta_L$ is related to  $\tilde{p}$ as given below.
\begin{eqnarray}
\tilde{p}&=&\frac{\exp(-\beta_L)}
                 {1+\exp(-\beta_L)}\nonumber\\
         & & \nonumber\\   
\beta_L &=& \log\Big( \frac{1}{\tilde{p}}-1\Big)
\end{eqnarray}
The procedure described above is repeated independently on all the pixels.
The {\it Likelihood} of $X\vert\widehat{\Theta}$ is then the product of the 
{\it Likelihoods} of $x_i\vert\widehat{\theta}_i$ and is given by,
\begin{eqnarray}
{\cal L}(X\vert\widehat{\Theta})&=&\frac{\exp\Big[-\beta_L F(X,\widehat{\Theta})\Big]}
                                        {\sum_X \exp\Big[-\beta_L F(X,\widehat{\Theta})\Big]},
\end{eqnarray}
where $F(X,\widehat{\Theta})$ is the Hamming distance between $X$ and $\widehat{\Theta}$.
The hyper-parameter $\beta_L=1/T_L$ can be interpreted as inverse temperature of the 
degradation process. 
When $T_L=0$ we have $\tilde{p}=0$. In other words 
at $T_L=0$ no degradation takes place and $X=\widehat{\Theta}$. 
As $T_L$ increases the noise level increases.   When
$T_L=\infty$, we have $\tilde{p}=0.5$ and  the image $X$ i
becomes  completely random. In other words each pixel 
can have independently
a black or white gray level with equal probability.

\section*{Degradation of a Multi-Gray Level Image}
\rhead[Bayesian Methodology]{\thepage}
\rhead[Markov Chain Monte Carlo Image Restoration]{\thepage}
Consider an image $\widehat{\Theta}$ painted with $Q$ gray levels with labels
$0,\ 1,\ 2,\ \cdots,\ Q-1$. The label $0$ denotes black and the label $Q-1$ denotes white.
Let us say that the degradation of $\widehat{\Theta}$ to $X$ proceeds as follows.
Take a pixel $i\in{\cal S}$. Call a random number $\xi$ uniformly distributed in the 
range $0$ to $1$. If $\xi\le \tilde{p}$, then select one of the $Q-1$ gray levels
(excluding $\widehat{\theta}_i$) randomly and with equal probability and assign it to $x_i$.
Otherwise set $x_i=\widehat{\theta}_i$. This process of degradation is described by the 
{\it Likelihood} given by,
\begin{eqnarray}
{\cal L}(x_i\vert\widehat{\theta}_i)&=&\tilde{p}\ {\cal I}(x_i\ne\widehat{\theta}_i)+
                    (1-\tilde{p})\ {\cal I}(x_i=\widehat{\theta}_i)
\end{eqnarray}
It is convenient to express the above {\it Likelihood} as,
\begin{eqnarray}
{\cal L}(x_i\vert\widehat{\theta}_i)&=&\frac{\exp\Big[-\beta_L {\cal I}(x_i\ne\widehat{\theta}_i\Big]}
                                            {1+ (Q-1)\exp(-\beta_L)}.
\end{eqnarray}
The hyper-parameter $\beta_L$ is related to  $\tilde{p}$ as given below,
\begin{eqnarray}
\tilde{p}&=&\frac{\exp(-\beta_L)}
                 {1+(Q-1)\exp(-\beta_L)}\nonumber\\
         & & \nonumber\\
\beta_L &=& \log\Big[ (Q-1)(\frac{1}{\tilde{p}}-1)\Big].
\end{eqnarray}
The degradation process described above is repeated independently on all the pixels.
Then the {\it Likelihood} of $X\vert\widehat{\Theta}$ is the product of the
{\it Likelihoods} of $x_i\vert\widehat{\theta}_i$ and is given by,
\begin{eqnarray}
{\cal L}(X\vert\widehat{\Theta})&=&\frac{\exp\Big[-\beta_L F(X,\widehat{\Theta})\Big]}
                                        {\sum_X \exp\Big[-\beta_L F(X,\widehat{\Theta})\Big]},
\end{eqnarray}
where $F(X,\widehat{\Theta})$ is the Hamming distance between $X$ and $\widehat{\Theta}$.
The hyper-parameter $\beta_L=1/T_L$ can be interpreted as inverse temperature of the
degradation process.
When $T_L=0$ we have $\tilde{p}=0$ indicating 
no degradation and $X=\widehat{\Theta}$.
As $T_L$ increases the noise level increases.   When
$T_L=\infty$, we have $\tilde{p}=1-(1/Q)$ which tends to unity when $Q\to\infty$.
In all the above if substitute $Q=2$ we recover the results of binary symmetric channel 
degradation. 

\section*{Three Ingredients of Bayesian Methodology}
In a  typical image processing algorithm, we start 
with an initial image $\Theta_0$. A good choice 
is $\Theta_0=X$, the  given noisy image. We generate a chain of images 
\begin{eqnarray}
\Theta_0\to\Theta_1\to\Theta_2\cdots\to\Theta_n\to\Theta_{n+1}\to
\end{eqnarray}
by an algorithm and hope that a suitably defined 
statistics over the ensemble of images in the chain 
would approximate well the original image $\widehat{\Theta}$.
In constructing a chain of images we shall employ Bayesian methodology.

The first ingredient of a Bayesian methodology is the {\it Likelihood} which 
models the degradation of $\widehat{\Theta}$ to $X$. 
We have already seen a few models of image degradation. 
The second ingredient  is an $\grave{{\rm a}}$ priori 
distribution, simply called a 
{\it Prior},  denoted by  $\pi(\Theta)$. The {\it Prior} 
is a distribution of gray levels on ${\cal S}$. The choice of a {\it Prior}
is somewhat subjective.  
The third ingredient  is an  $\grave{{\rm a}}$ 
posteriori distribution or simply called 
 {\it Posterior}, denoted by the symbol $\pi(\Theta\vert X)$. 
{\it Posterior} is a conditional distribution; 
it is the distribution of $\Theta\vert X$.

\subsection*{Bayesian $\grave{{\rm{\bf a}}}$ Priori Distribution}
As we said earlier, the choice of a {\it Prior} in Bayesian methodology 
is subjective. The {\it Prior}
should reflect what we 
believe a clean image should look like.  We expect  an image 
to  be smooth. We say that a pixel is smoothly connected to its neighbour 
if the 
gray levels in them are the same. 
Smaller the difference between the gray levels, smoother is the connection.
This subjective belief is encoded in a {\it Prior} as follows. 

We introduce an interaction energy  between the states of two pixels.
We confine the interaction to nearest neighbour pixels. Thus the states of a pair $i,j$ of nearest 
neighbour pixels interact with an  energy  denoted by $E_{i,j}$.
We take the interactions to be pair-wise additive and  define an energy 
of the image $\Theta$ as,
\begin{eqnarray}
E(\Theta)&=&\sum_{\langle i,j\rangle} E_{i,j}(\Theta) .
\end{eqnarray}
In the above, the symbol $\langle i,j\rangle$ denotes
that the pixels $i$ and $j$
are nearest neighbours and the sum is taken over all distinct 
pairs of nearest neighbour
pixels in the image plane. We model $E_{i,j}$ in such a way that 
it is small when the gray levels are close to each other and  large when the 
gray levels differ by a large amount.  Thus, a smooth image 
has less energy. We define the {\it Prior} as,
\begin{eqnarray}
\pi(\Theta)&=&\frac{1}{Z(\beta)}\exp\Big[ -\beta_P E(\Theta)\Big]
\end{eqnarray}  
where $\beta_P=1/T_P$ and $T_P$ is a smoothening parameter called the {\it Prior} 
temperature. $Z(\beta_P)$ 
is the normalization constant and is given by,
\begin{eqnarray}\label{Prior_Z}
Z(\beta_P)&=&\sum_{\Theta}\exp\Big[-\beta_P E(\Theta)\Big].
\end{eqnarray}
The {\it Prior}, as defined above, is called the Gibbs distribution. 
Such a definition of {\it Prior\  }  implies that $\Theta$ is a Markov random field. Conversely if 
$\Theta$ is a Markov random field, then its {\it Prior} 
can be modeled as a   Gibbs distribution. 

\section*{Markov Random Field}
\rhead[Ising Prior]{\thepage}
\rhead[Markov Chain Monte Carlo Image Restoration]{\thepage}
A Markov Random field generalizes the notion of a discrete time Markov process,
or also called a Markov chain.  Appendix (2) briefly describes 
a general as well as a time homogeneous or equilibrium
Markov chain. A sequence of random variables, 
parametrized by time and obeying 
Markovian  dependence, constitutes  a Markov chain. Instead of time
we can use  spatial coordinate as  parameter to describe the chain,
in one dimensional setting. The concept of Markov random field 
extends Markovian dependence from one dimension to a general setting
\cite{RLD}

In a Markov random field $\Theta$, the state  of a pixel is at best dependent
on the state  of the pixels in its neighbourhood\footnote{Often the nearest neighbours of a pixel 
constitute its neighbourhood; for example
the four pixels situated to the  left, right, 
above and below the pixel $i$ form the 
neighbourhood of $i$; this definition of neighbourhood 
is employed in the square lattice
models of statistical mechanics. In image processing, 
the eight pixels - four nearest neighbours and  four
next-nearest neighbours,
are taken as neighbourhood in several applications,  
like for example 
low pass filter. Even $24$ pixels surrounding a central pixel are 
also often considered as neighbourhood, in filter algorithms.}. 
Accordingly, let $i\in{\cal S}$ be a pixel and 
let $\nu_i$ denote the set of pixels that form 
the neighbourhood  of $i$. Let ${\cal N}=\{ \nu_i\ :\  i\in{\cal S}\}$ 
denote a neighbourhood system for ${\cal S}$. In other words,
${\cal N}$ is any collection of subsets of ${\cal S}$ for which,
\begin{enumerate}
\item[1)] $i\notin\nu_i$
\item[2)] $i\in\nu_j \Longleftrightarrow j\in\nu_i$.
\end{enumerate} 
The pair $\{ {\cal S}, {\cal N}\}$ is a graph. 
An image $\Theta$ is a Markov random field if
\begin{eqnarray}
\pi\Big(\theta_i\Big\vert\{\theta_j\ :\ j\ne i;j\in{\cal S}\}\Big)=
\pi\Big(\theta_i\Big\vert\{\theta_j\ :\ j\in\nu_i\}\Big)\ \forall\ i\in{\cal S}.
\end{eqnarray}
From the definition of conditional probability in terms of joint probability, 
we have,
\begin{eqnarray}
\pi (\Theta)&=&\pi(\theta_1,\theta_2,\cdots ,\theta_N)\nonumber\\
            & & \nonumber\\
            &=&
\pi\Big(\theta_i\Big\vert\big\{ \theta_j : j\ne i; j\in {\cal S}\big\}\Big)
\pi(\theta_j :j\ne i; j\in{\cal S})\ \ \forall\ \ i\in\ {\cal S}
\end{eqnarray}
Since $\Theta$ is a Markov random field, the above can be written as,
\begin{eqnarray}
\pi(\Theta)=\pi\Big( \theta_i\Big\vert\big\{ \theta_j : j\in\nu_i\big\}\Big)
\pi(\theta_j :j\ne i; j\in{\cal S})\ \forall\ i\in{\cal S}.
\end{eqnarray}
Iterating alternately, we get,
\begin{eqnarray}
\pi (\Theta) = \prod_{i\in{\cal S}}\ \pi\Big( \theta_i\Big\vert\big\{ \theta_j :j\in\nu_i\big\}\Big).
\end{eqnarray}
Let us define, 
\begin{eqnarray}
U_i(\Theta)=-\log\pi \Big( \theta_i\Big\vert\{\theta_j : j\in\nu_i\}\Big),
\end{eqnarray}
and
\begin{eqnarray}
U(\Theta)=\sum_{i\in{\cal S}} U_i (\Theta).
\end{eqnarray}
Then,
\begin{eqnarray}
\pi(\Theta)=\frac{1}{Z}\exp\big[ - U(\Theta)\big],
\end{eqnarray} 
where $Z$ is the normalization constant. This expression for
the {\it Prior} is identical to the Gibbs distribution if we identify
$U(\Theta)=\beta_P E(\Theta)$ and $Z$ as the 
$Z(\beta_P)$, defined in Eq. (\ref{Prior_Z}).
Note that $ E(\Theta)$ is a sum of the interaction energies 
of the states of nearest neighbour pixels and hence
is consistent with the Markov random field requirement.
It is important that the interaction must be restricted 
to a  finite range  for
$\Theta$ to be a Markov random field. Since 
we consider in this review  only nearest neighbour
interactions, this condition is automatically satisfied.

The equivalence of Gibbs distribution in 
statistical mechanics and Markov random field
in spatial statistics was established, see \cite{JEB_1974,OFDS}  
following the work of  Hammersley and Clifford \cite{JMHPC}.

\section*{Ising Model for $\grave{{\rm {\bf a}}}$  Priori Distribution}
Ising model \cite{Ising} is the simplest and 
perhaps the most studied of models in statistical mechanics.
Let us consider a {\it Prior} inspired by Ising model. 
We consider an image  having 
only two gray levels,  designated as say $\zeta=1,\ 3$ or equivalently 
$S=\pm 1$; we have the transformation $\zeta=S+2$.   
The index $\zeta=1$ ($S=-1$) refers to black and 
$\zeta=2$ ($S=+1$) refers to white gray level. The interaction energy 
of the states of two nearest neighbour pixels $i$ and $j$ is given by,
\begin{eqnarray}
E_{i,j}(\Theta)&=&  {\cal I}(\theta_i\ne\theta_j).
\end{eqnarray}
It is clear that if the two 
pixels have the same gray level, the interaction energy 
is zero. If their gray levels  are 
different the interaction energy is unity. 
In the language of the physicists, the ground state 
of the two-pixel system has zero energy and the 
excited state has unit energy.  The two energy levels
are separated by $\Delta E=1$. The energy of an Ising  image 
$\Theta$ is given by\footnote{The Ising Hamiltonian in statistical mechanics 
is given by $E=-J\sum_{\langle i,j\rangle}S_i S_j$,
where $S_i=\pm1$ is the spin on the lattice site $i$ 
and $J$ is the strength of spin-spin interaction,
usually set to unity. The ground state of the a 
pair of nearest neighbour spins is of energy $-1$ and the excited
state is of energy $+1$; the energy separation, 
is thus $2$; {\it i.e.} $\Delta E=2$},
\begin{eqnarray}
 E(\Theta)=\sum_{\langle i,j\rangle }
E_{i,j}(\Theta)=\sum_{\langle i,j\rangle }
{\cal I}\left(\theta_i\ne\theta_j\right).
\end{eqnarray}
The Ising {\it Prior} can be written as,
\begin{eqnarray}\label{Ising-Potts_Prior}
\pi(\Theta)&=&\frac{1}{Z(\beta_P)}
\exp\Big[-\beta_P
\sum_{\langle i,j\rangle}{\cal I}(\theta_i\ne\theta_j)\Big]
\end{eqnarray}
where $Z(\beta_P)$ is the canonical partition function. 
Ising {\it Prior} in conjunction with Hamming  {\it Likelihood} is 
suitable for restoring binary  images. 

\section*{Potts Model for  $\grave{{\rm {\bf a}}}$ Priori Distribution}
Potts model \cite{Potts} is an important lattice model in which there 
are discrete number of states at each site.
Consider a multiple gray level image. 
The gray levels are taken as $\{ 0, \ 1,\ 2,\ \cdots ,Q-1\}$.
We have then a $Q$ state Potts model. The label  
$0$ denotes black, and the label $Q-1$ 
denotes white gray levels. 
The different shades correspond to the intermediate 
Potts numbers.  
The expressions for the energy associated with the 
states of two nearest neighbour pixels,  for the 
Potts Hamiltonian and for the Potts {\it Prior} are the same 
as the ones defined for the Ising model given 
in the last section\footnote{In statistical mechanics, 
the energy associated with a pair of nearest neighbour
Potts spins $-J\delta_{S_i,S_j}$, where $S_i=0,1,\cdots , Q-1$ are the Potts spin labels and 
$J$ is the strength of spin-spin interaction, usually set to unity. The ground state is 
of energy zero and the excited state is of energy unity; the energy separation is unity.
{\it i.e.} $\Delta E=1$.}. 
Potts {\it Prior} in conjunction with 
Hamming  or  Kullback-Leibler  {\it Likelihood} 
is suitable for restoring images painted 
with a  small number of gray levels. 

\section*{Gemen-McClure Model for $\grave{{\rm {\bf a}}}$  Priori Distribution}
\rhead[Relation between $\beta_P$ and $\widehat{\Theta}$]{\thepage}
\rhead[Markov Chain Monte Carlo Image Restoration]{\thepage}
Gemen and McClure\cite{SGDEM_1987} 
recommended  a {\it Prior} in which the states of two neighbouring 
pixels interact with an energy given by,
\begin{eqnarray}
E_{i,j} (\Theta)=-\frac{1}{1+C(\theta_i -\theta_j)^2},
\end{eqnarray}
where $C$ is a hyper-parameter that determines the width of the distribution. 
The value of $E_{i,j}(\Theta)$ ranges from  a minimum of $-1$ 
when $\vert\theta_i-\theta_j\vert =0$ to a maximum of $0$, 
when $\vert\theta_i-\theta_j\vert\to\infty$. Gemen-McClure interaction potential  is 
depicted in Fig.\ (2)
for $C=0.1,\ 1.0,\ {\rm and}\  10.0$.
The function $E_{i,j}(\Theta)$ is symmetric
about  $\vert\theta_i-\theta_j\vert=0$ and its width decreases with increase of
$C$. 
For large $C$ the Gemen-McClure interaction  reduces to  an Ising interaction  with 
ground state energy at $-1$ and excited state energy at $0$. 
The Gemen-McClure {\it Prior} is thus given by,
\begin{eqnarray}
\pi(\Theta)=\frac{1}{Z(\beta_P)}\ \exp\bigg[\ -\beta_P\ \sum_{\langle i,j\rangle}
\ \frac{-1}{1+C\ (\ \theta_i -\theta_j\ )^2\ }
\bigg]
\end{eqnarray}
where $Z(\beta_P)$ is the partition function that normalizes the  {\it Prior}.
\begin{figure}[pt]
\label{Gemen-McClure-Energy_fig}
\centerline{\includegraphics[height=3.9in,width=5.5in]{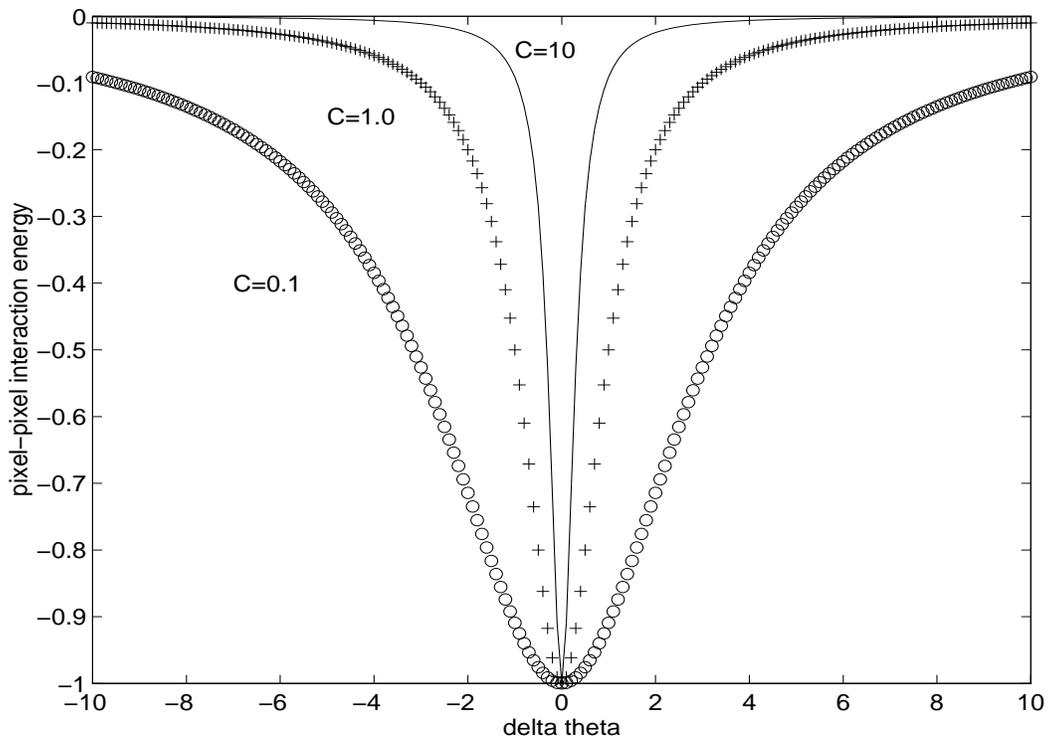}}
\caption{ \small {\bf Gemen-McClure energy $E_{i,j}$ versus 
$\vert\theta_i=\theta_j\vert$ for $C=0.1,1.0,10.0$; when $C$ increases the 
width of the function decreases. In the limit $C\to\infty$ the Gemen-McClure
model reduces to the Ising model with ground state at $-1$ and excited state at $0$.}}
\end{figure}

\subsection*{Relation between $\beta_P$ and $\widehat{\Theta}$}
It is clear from the discussion above that the hyper-parameter $\beta_P$ 
will depend on the smoothness and features present in $\widehat{\Theta}$.
The {\it Prior} is given by,
\begin{eqnarray}
\pi(\widehat{\Theta})&=&\frac{1}{Z(\beta_P)}\exp\Big[ -\beta_P E(\widehat{\Theta})\Big],
\end{eqnarray}
where $Z(\beta_P)$ is the partition function,
\begin{eqnarray}
Z(\beta_P)=\sum_{\widehat{\Theta}}\exp\Big[ -\beta_P E(\widehat{\Theta})\Big]
\end{eqnarray}
The value of $\beta_P$ is implicitly given by the relation below.
\begin{eqnarray}
\langle E\rangle &=& -\frac{\partial}{\partial{\beta_P}}\log\Big[ Z(\beta_P)\Big] 
\end{eqnarray}
Consider for example the Ising/Potts {\it Priors}. 
The energy of $\widehat{\Theta}$ is completely determined 
by the number of pairs of nearest neighbour pixels with dis-similar 
gray levels\footnote{Two nearest neighbour pixels having the same gray level are called
similar pixels. If the gray levels are different they are called dissimilar pixels}.
In the graph theoretic language two nearest neighbour pixels are separated 
by an edge. Let $N_E$ denote the total number of edges in image plane of 
$\widehat{\Theta}$. If two nearest neighbour pixels have the same gray levels
then we say the edge separating them is a satisfied 
bond\footnote{this terminology will be useful later when we consider cluster algorithms
for image restoration}.
Let $B(\widehat{\Theta})$ denote the number of satisfied bonds in the image $\widehat{\Theta}$.
In other words $B$ is the number of nearest neighbour pairs of similar pixels. 
We see immediately that $E(\widehat{\Theta})=N_E-B$. Thus the value of the hyper-parameter
$\beta_L$ can in principle be calculated from the values of $N_E$ and $B$ of a given 
image $\widehat{\Theta}$.  
 
\section*{Bayesian $\grave{{\rm{\bf a}}}$  Posteriori Distribution}
\rhead[MAP estimate]{\thepage}
\rhead[Markov Chain Monte Carlo Image Restoration]{\thepage}
The {\it Posterior} distribution in the Bayesian methodology 
is given by the product of the 
{\it Likelihood} and the {\it Prior}. This is called Bayes' theorem, 
discussed for {\it e.g.} in
\cite{Feller,Papoulis}. Appendix 1 states
Bayes' theorem.

According to Bayes' theorem,
\begin{eqnarray}
\pi(\Theta\vert X)=
\frac{{\cal L}(X\vert\theta)\pi(\theta)}
{\sum_{\Theta}{\cal L}(X\vert\theta)\pi(\theta)}.
\end{eqnarray}
The Bayesian {\it Posterior} can be formally written as,
\begin{eqnarray}
\pi(\Theta\vert X)=
\frac{\exp\bigg[ -\bigg\{ \beta_L F(\Theta,X) +
\beta_P E(\Theta)\bigg\}\bigg]}
                        {\sum_{\Theta}
\exp\bigg[ -\bigg\{ \beta_L F(\Theta,X) +
\beta_P  E(\Theta)\bigg\}\bigg]}.
\end{eqnarray}
It is convenient to work in terms of intensive quantities. 
To this end we define $E$ per pair of nearest 
neighbour pixels and $F$  per pixel. Accordingly  we divide  
energy by, $N_E$, the number of nearest neighbour pairs of 
pixels in the image plane. For an $L\times L$
pixel image, we have $N_E=2L(L-1)\approx 2L^2 ({\rm for\  large\ } L)$. 
We divide 
$F(\Theta,X)$ by $L^2$.
In the expression for the {\it Posterior} above, 
the argument of the  exponential function can be viewed as  a weighted sum of 
$F(\Theta,X)$ and $ E(\Theta)$.  
The weight attached to $E$ is $\beta_P$ and the weight 
attached to the $F$ is $\beta_L$. The relevant quantity is the relative weights 
attached to $F$ and $E$. Hence we set $\beta_L=1$ and  $\beta_P=\beta$, where $\beta$ is the 
value of $\beta_P$ measured in units of $\beta_L$. Also, it is convenient to work with 
normalized weights and accordingly we divide the whole exponent by
$1+\beta$. The final expression for the {\it Posterior} that we use in the image restoration 
algorithm is given by
\begin{eqnarray}
\pi(\Theta\vert X)=
\frac{\exp\bigg[ -L^{-2}(1+\beta)^{-1}\Big\{ F(\Theta,X) +
2^{-1}\beta  E(\Theta)\Big\}\bigg]}
                        {\sum_{\Theta}
\exp\bigg[ -L^{-2}(1+\beta)^{-1}\Big\{ F(\Theta,X) +
2^{-1}\beta  E(\Theta)\Big\}\bigg]}.
\end{eqnarray}
The advantage of the above expression is there is only one 
parameter $\beta=1/T$ that needs to be tuned for good image 
restoration\footnote{It is purely for convenience that we have reduced the 
number of hyper-parameters from two to one. Thus we have only one 
hyper-parameter denoted by $\beta=1/T$. We call $T$ the temperature. 
For many applications we would require two or more hyper-parameters 
which need  to be optimized
for good image restoration. Pyrce and Bruce \cite{JMPADB}, for example,
have considered both  $\beta_L$ and $\beta_P$ separately and 
carried out Monte Carlo search in the two dimensional hyper-parameter space for
good image restoration. In fact they talk of first order transition line that  
separates the {\it Prior} dominated  and the 
{\it Likelihood} dominated phases in the context of Ising spin model 
of image restoration.}.
 
\section*{Bayesian Maximum $\grave{{\rm{\bf a}}}$ Posteriori (MAP)}
The aim of image restoration is to construct 
the true image $\widehat{\Theta}$ from the given noisy
image $X$. The simplest estimate of 
$\widehat{\Theta}$ is the image which maximizes the {\it Posterior}. 

For a given temperature, the {\it Posterior}
increases when $F(\Theta , X)$ decreases. In other words
closer  $\Theta$ is to $X$,  larger is the 
value of $\pi(\Theta\vert X)$.
Though $X$ is  corrupt,
it is the best we have and
we would like to retain during  image-restoration
process as many  features of  $X$ as possible.

The {\it Posterior} increases when $E$ 
decreases for a given temperature $T$. In other words
smoother  $\Theta$ is,  larger is the value 
of  $\pi(\Theta\vert X)$.
Thus, there is a {\it Prior}-{\it Likelihood} competition, 
the {\it Prior} trying to smoothen the image and the 
{\it Likelihood} trying to keep the image as close to $X$ 
as possible. In other words the {\it Prior} tries to make all the pixels 
acquire the same grave level and  the {\it Likelihood} tries to bind the image
to the data $X$. As a result we get an image
which has in it the features of $X$ 
(because of  competition from the {\it Likelihood}) and which is smooth 
with  out noise (because of competition from the 
{\it Prior})\footnote{We can say that this kind of 
{\it Prior}-{\it Likelihood} competition
is  analogous to the entropy-energy competition 
which determines the phase of a material.}.
The nature of {\it Prior}-{\it Likelihood} competition 
is tuned by  Temperature, see below. 
\vskip 3mm
\noindent
What happens when the temperature $T$ is large ? 
\vskip 3mm
\noindent
Temperature 
determines  the relative competitiveness of the {\it Likelihood} and {\it Prior} toward  increasing the {\it Posterior}.
For a given $T$,

\begin{enumerate}
\item[-]the relative weight attached to $F(\Theta,X)$
is $T/(1+T)$, and
\item[-]
the relative weight attached to $E$ is
 $1/(1+T)$.
\end{enumerate}
When $T$ is large,  {\it feature - retention} ({\it i.e.} the {\it Likelihood})
is given more importance.
However if $T$ is very large, the image restoration algorithm would interpret even the
noise in $X$ as  a feature and retain it;
as a result the restored image would  be noisy.
\vskip 3mm
\noindent
What happens when the temperature $T$ is small ?
\vskip 3mm
\noindent
When $T$ is small, smoothening ({\it i.e.} the {\it Prior}) is given more importance.
However if $T$ is too small, there is a  danger:  the image restoration algorithm
would misinterpret even genuine inhomogeneities and fine features of $X$ as
noise and eliminate them.

Best image restoration is expected over an intermediate
range of $T$, often  obtained by trial and error.
The problem of image restoration  is thus reduced to a problem
of sampling, independently and randomly with equal probabilities, 
a large number  of images from the state space $\Omega$.  These images  constitute  
a microcanonical ensemble. Calculate the {\it Posterior} of each member 
of the microcanonical ensemble.  Pick up the image which  gives 
the highest value of the  {\it Posterior} and call it $\Theta_{{\rm MAP}}$; the 
suffix MAP stands for the Maximum $\grave{{\rm a}}$ Posteriori. $\Theta_{{\rm MAP}}$ is
called an MAP estimate of 
$\widehat{\Theta}$. We can also employ other statistics,  defined over the state space $\Omega$,
to estimate $\widehat{\Theta}$,
see below. 

\section*{Maximum $\grave{{\rm{\bf a}}}$ Posteriori Marginal (MPM) 
}
\rhead[Markov Chain Monte Carlo Image Restoration]{\thepage}
We partition the state space $\Omega$ into
mutually exclusive and exhaustive subsets  as described below.
Consider a pixel $i\in{\cal S}$. Define $\Omega^{(i)}_{\zeta}$ as a subset of 
images for which the gray level of the pixel $i$ is $\zeta$, where 
$\zeta\in\{ 0,\ 1,\ \cdots Q-1\}$. 
\begin{eqnarray}
\Omega^{(i)}_{\zeta}=\bigg\{ \Theta\in\Omega ,\  \theta_i(\Theta)=\zeta\bigg\}\ 
\ {\rm for}\  \ \zeta = 0,1,\cdots Q-1
\end{eqnarray}
Calculate now marginal {\it Posteriors},
\begin{eqnarray}
\pi^{(i)}_{\zeta}=\sum_{\Theta\ \in\ \Omega^{(i)}_{\zeta}}\pi(\Theta\vert X)
\ \ \ {\rm for}\ \ \ \zeta=0,1,\cdots Q-1\ .
\end{eqnarray}
Thus we get an array of $Q$ marginals 
$\Big\{ \pi^{(i)}_{\zeta}:\zeta=0,1,\cdots Q-1\Big\}$.
We define
\begin{eqnarray}
\zeta^{(i)}_{{\rm MPM}}&=&{\rm arg} \begin{subarray}
                                           \ \cr
                                           \ \cr
                                          \ {\rm{max}} \cr
                                           \ \ \zeta
                                          \end{subarray}
                                            \ \ \pi^{(i)}_{\zeta}
\end{eqnarray}
which stands for the value of $\zeta$ that maximizes 
the function $\pi^{(i)}_{\zeta}$.  
In other words, $\zeta^{(i)}_{{\rm MPM}}$ denotes 
the value of the gray level for which 
the marginal {\it Posterior} is maximum.
Repeat the above for all the pixels in the image plane, 
and obtain a collection of numbers that represents 
an approximation to $\widehat{\Theta}$:
\begin{eqnarray}
\Theta_{{\rm MPM}}=\Big\{ \zeta^{(i)}_{{\rm MPM}} : i\in {\cal S} \Big\}.
\end{eqnarray}
$\Theta_{{\rm MPM}}$ is called a Maximum $\grave{{\rm a}}$ Posteriori (MPM) 
estimate of  $\widehat{\Theta}$. 

\section*{Threshold $\grave{{\rm{\bf a}}}$ Posteriori Mean (TPM) }
\rhead[Elements of Image Processing]{\thepage}
\rhead[Markov Chain Monte Carlo Image Restoration]{\thepage}
Calculate an average image, $\overline{\Theta}=\{ \overline{\theta}_i\ :\ i\in{\cal S}\}$, where,
\begin{eqnarray}
\overline{\theta_i}=\sum_{\Theta\in\Omega}\theta_i(\Theta)\ \pi(\Theta\vert X)\ \ \forall\ \ i\ \ \in\ \ {\cal S}.
\end{eqnarray}
We define,
\begin{eqnarray}
\zeta^{(i)}_{{\rm TPM}}&=&{\rm arg} \begin{subarray}
                                           \ \cr
                                           \ \cr
                                          \ {\rm{ min}} \cr
                                           \ \ \zeta
                                          \end{subarray}
                                            \ \ \Big[\zeta-
                                   \sum_{\Theta\in\Omega}\theta_i(\Theta)\pi(\Theta\vert X\Big]^2
\end{eqnarray}
which stands for the value of $\zeta$ that 
minimizes $[\zeta-\sum_{\Theta\in\Omega}\theta_i(\Theta)]^2$.
In other words $\zeta^{(i)}_{{\rm TPM}}$ 
is the value of $\zeta$ closest to $\overline{\theta_i}$. 
Carry out the above exercise for all the pixels in the image,
and get a collection of gray levels that represents an approximation
to $\widehat{\Theta}$: 
\begin{eqnarray}
\Theta_{{\rm TPM}}=\Big\{ \zeta^{(i)}_{{\rm TPM}} : i\in {\cal S} \Big\}.
\end{eqnarray}
$\Theta_{{\rm TPM}}$ is called Threshold $\grave{{\rm a}}$ Posteriori Mean (TPM)  
estimate of  $\widehat{\Theta}$. 
It is easily seen that for a binary image 
$\Theta_{{\rm TPM}}=\Theta_{{MPM}}$. 

\section*{Elements of Digital Image Restoration}
A typical  image restoration process is depicted  in Fig. (3).
We represent the unknown true image by $\widehat{\Theta}$, 
which gets corrupted
to $X$. The degradation of $\widehat{\Theta}$ to  $X$ is modeled by the 
{\it Likelihood} ${\cal L}(X\vert\widehat{\Theta})$. Bayes theorem
helps synthesize the Likelihood  (available in the form of 
a degradation model and data on the corrupt image $X$) with a 
{\it Prior } (that models  
your subjective beliefs about true image).  The result is a  
{\it Posterior} $\pi(\Theta\vert X)$. An ensemble of images
having the {\it Posterior } distribution is shown as $\Theta$ in 
Fig. (3). From this {\it Posterior ensemble } we can make
MAP, MPM and TPM estimates of the true image. 
\begin{figure}[htp]
\centerline{\includegraphics[height=1.8in,width=5.3in]{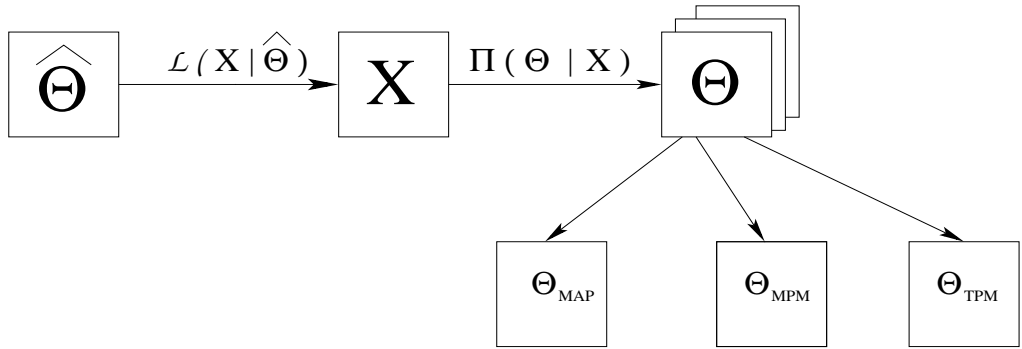}}
\caption{\small{\bf 
Digital image restoration : 
$\widehat{\Theta}$ denotes the 
true image (first square in the top row) 
which gets corrupted to $X$ (second square in the top row). 
The degradation of 
$\widehat{\Theta}$ to $X$ is modeled by ${\cal L}(X\vert\widehat{\Theta})$.
Bayes theorem helps construct a {\it Posterior } $\pi(\Theta\vert X)$
from the {\it Likelihood} 
and a {\it Prior}. A {\it Posterior} ensemble of images ( 
set of squares in the right end of top row). MAP, MPM and TPM estimates
of $\widehat{\Theta}$ (depicted by the three squares in the second row) 
are made from the {\it Posterior} ensemble. See \cite{JMPADB}}}
\end{figure} 
\section*{Algorithm for Calculating 
${\mathbf \Theta_{{\rm{\textbf MAP}}}}$}
A straight forward procedure  to calculate $\Theta_{{\rm MAP}}$  
is to attach a {\it Posterior} probability to each image belonging to the 
state space $\Omega$ and pick up that image which 
has a maximum {\it Posterior} probability. This is easily said than done. 
Note that the number of images belonging to state space 
$\Omega$ is $Q^N$. Let us consider 
a small, say $10\times 10$ binary image ($Q=2$). 
The number of pixels is thus $100$. 
Then $\widehat{\Omega}=2^{100}\approx 10^{30}$. 
Calculating the {\it Posteriors} for these 
$10^{30}$ images is an impossible task even on the 
present day high speed computers\footnote{A typical image 
is  of size between $256\times 256$ to $1024\times 1024$ with 
gray levels ranging from $2$, for a binary image to $256$. 
The state space will  contain order of $10^{10,000}$ images.}.
A simple  procedure would be to
sample randomly and with equal probability a certain large number of  images in the neighbourhood
of the given image $X$; these images constitute a microcanonical 
ensemble. 
Find the image, in the microcanonical ensemble, 
that maximizes the {\it Posterior} and recommend it as an MAP estimate of 
$\widehat{\Theta}$. An algorithm  for doing this is described below. 

Fix the temperature. Start with an image $\Theta_0=X$.
Call this the current image. {\it i.e.} $\Theta_C=\Theta_0$
\begin{enumerate}
\item[(1)]
Calculate $\pi_C =\pi(\Theta_c\vert X)$. 
\item[(2)]
Select a pixel, say $j$,
randomly from the image plane. 
\item[(3)]
Switch the gray level of the pixel $j$
from its current value to a value  
sampled randomly and with equal probability
from the gray level labels\\ $\zeta=\{ 0,\ 1,\ \cdots~,\ Q-1\}$. 
\item[(4)]
Calculate $\pi_t=\pi(\Theta_t\vert X)$. 
\item[(5)]
If $\pi_t > \pi_C$, then accept the 
trial image and set $\Theta_1=\Theta_t$; otherwise $\Theta_1=\Theta_0$.
\item[(6)]
Take $\Theta_1$  as the current image $\Theta_C$, 
\item[(7)] go to step (1).
\end{enumerate}
Iterate the whole process 
several times.
A set of $N$ update attempts constitutes an iteration. 
Collect the images at the end of each iteration. Thus we get  
a sequence of images with monotonically non-decreasing {\it Posteriors}.
Asymptotically we get $\Theta_{{\rm MAP}}$, called an  MAP estimate of 
$\widehat{\Theta}$. 

\subsection*{How does Posterior maximization lead to image restoration~?}
\rhead[Restoration of Toy and Benchmark Images]{\thepage}
\rhead[Markov Chain Monte Carlo Image Restoration]{\thepage}
A simple explanation of how does a Bayesian Posterior maximization 
lead to de-noising is depicted in Fig. (4) and described below. 
\begin{figure}[htp]
\centerline{\includegraphics[height=2.0in,width=6.0in]{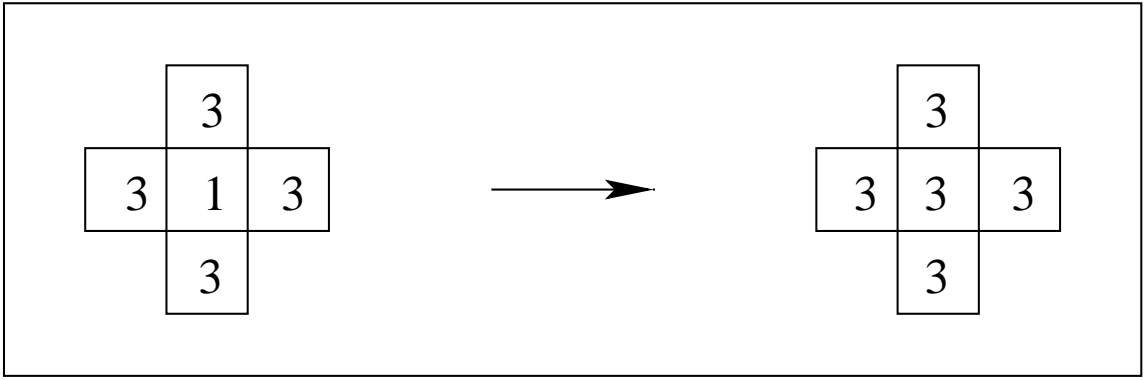}}
\caption{\small {\bf Local region of $\Theta_C$ is depicted at Left.
The gray level label of the central pixel is switched from $1$ to $3$ and 
a trial image $\Theta_t$ is constructed. The corresponding local region of 
$\widehat{\Theta}_t$ is depicted at Right.  This move leads to local 
smoothening of the image.
Let us assume that the sub-image depicted at Left belongs to $X$
and that depicted at Right belongs to $\widehat{\Theta}$. 
Then the above  move increases  
$F$ by one unit  and  decreased  $ E$ by $4$ units. 
The {\it Posterior}
increases if $T < 2$. In other words {\it Prior} wins over 
{\it Likelihood} if $T_L \ <\ 2$, leading to de-noising. }}
\end{figure}
Consider a local region of the 
current image $\Theta_C$ having a pixel $m$ with gray level 
label $1$ and with all its four nearest neighbours having gray level 
label $3$. Let us denote by the symbol $\nu$ the set of five 
pixels : the pixel $m$ and its four nearest neighbours.
Let us say that in $\widehat{\Theta}$, the pixels belonging to
$\nu$ have all the same gray level $3$ and hence is smooth locally.
It is the process of  degradation 
that has to led to the noise inhomogeneity in the  pixel $m$.
We have,
\begin{eqnarray}
F(\Theta_C, X) &=& \sum_{i\notin\nu}
{\cal I}\Big( \theta_i(\Theta_C)\ne x_i\Big)
+\sum_{i\in\nu}{\cal I}
\Big( \theta_i(\Theta_C)\ne x_i\Big)
\end{eqnarray}
Let us denote the first term in the above sum (over pixels not belonging
to $\nu$) as $F_0$. 
The second term is zero since $\theta_i(\Theta_C)=x_i(X)\ \forall\ i\in\nu$. 
We get $F(\Theta_C, X)=F_0$. Similarly,
\begin{eqnarray}
E(\Theta_C)&=&\sum_{\begin{subarray}
                             \ \langle i,j\rangle\cr
                               i\ne m\end{subarray}
                          }{\cal I}\Big( \theta_i(\Theta_C)\ne\theta_j(\Theta_C)
                                   \Big)
                          + \sum_{\begin{subarray}
                           \ \langle i,j\rangle\cr
                             i=m\end{subarray}}
                           {\cal I}\Big( \theta_i(\Theta_C)\ne\theta_j(\Theta_C)
                                    \Big),
\end{eqnarray}
where $\langle i,j\rangle$ denotes that $i$ and $j$ are 
nearest neighbour pixels and the sum extends over all 
distinct nearest neighbour pairs of pixels. 
Let us denote by $ E_0$ the first term in the above.
The second term is $4$ since there are four dis-similar 
 nearest neighbour pairs of pixels in the 
sub-image $\nu$. Hence $E(\Theta_C)=E_0 + 4$. 
In the simulation, we change the gray level of pixel $m$ from it 
present value of $1$ to $3$ and call the resulting image as 
$\Theta_t$. We find that $F(\Theta_t,X)=F_0 +1$ and $E(\Theta_t)=E_0.$ 
Eventhough the move increases $F$ by one unit, it decreases $E$
by four units. Let the {\it Posteriors} of $\Theta_C$ and $\Theta_t$ 
be denoted by $\pi_C$ and $\pi_t$ respectively. The ratio of the 
{\it Posteriors} can be calculated and is given by,
\begin{eqnarray}
\frac{\pi_t}{\pi_C}=\exp\bigg[ -\frac{1}{L^2(1+\beta)}\ \Big( 1-2\beta
\Big)\bigg]
\end{eqnarray} 
which is greater than unity whenever $\beta\ >\ (1/2)$ or
$T\ <\ 2$. Since we accept $\Theta_t$  
only when $\pi_t$ is greater than $\pi_C$, 
de-noising takes place when $T <  2$.

It is clear from the discussion above, 
that removal of an inhomogeneity entails energy reduction. Hence a {\it Prior} tries to remove
all inhomogeneities in the image plane including the genuine ones. On the other hand,
the {\it Likelihood} binds the image to the data $X$. It tries to retain all the 
features of $X$ including the inhomogeneities that have their origin in  noise. It is 
the temperature that tunes the competition between the {\it Prior} and the 
{\it Likelihood}.  

\section*{Restoration of Toy Images}
\subsection*{Binary Robot}
We have created  a $92\times 92$  binary image $\widehat{\Theta}$, 
referred to as ROBOT, depicted in 
Fig.~(5~:~Left). To corrupt a binary image with noise we proceed as follows.
We select a pixel $i$ 
and change the gray level label from its present value to the 
other with a probability $0.05$: Call a random number $\xi$; if 
$\xi\le 0.05$, change the gray level. Otherwise do not change the gray level.
Carry out this exercise on all the pixels independently.
This is equivalent to adding $5\%$ noise to the 
image; there is no spatial correlations in the noise added. 
The resulting corrupt image, called $X$ is depicted in 
Fig.~(5~:~Middle). Employing Ising {\it Prior} and Hamming 
{\it Likelihood} we have calculated an MAP estimate of 
$\widehat{\Theta}$ at $T=0.51$ depicted in Fig.~(5~:~Right). 
\subsection*{Five-gray Level Robot}
We have constructed a $5$ gray level ROBOT image on a $56\times 56$ 
image frame depicted in Fig. (6 : Left). We corrupt the image 
with $5\%$ noise and get $X$ depicted in Fig. (6 : Middle).
Employing Potts {\it Prior} and Hamming {\it Likelihood} we 
have made an MAP estimate of the true image which is depicted in 
Fig. (6 : Right). The image restoration has been carried out at 
$T=0.51$. 
We monitored the Bayesian {\it Posterior} maximum at the end of 
each iteration. The {\it Posterior} maximum is a monotonically
non-decreasing of the iteration index. Eventually the {\it Posterior}
maximum saturates as shown in Fig. (7).
\begin{figure}[ht]
\centerline{\includegraphics[height=2.0in,width=6.0in]{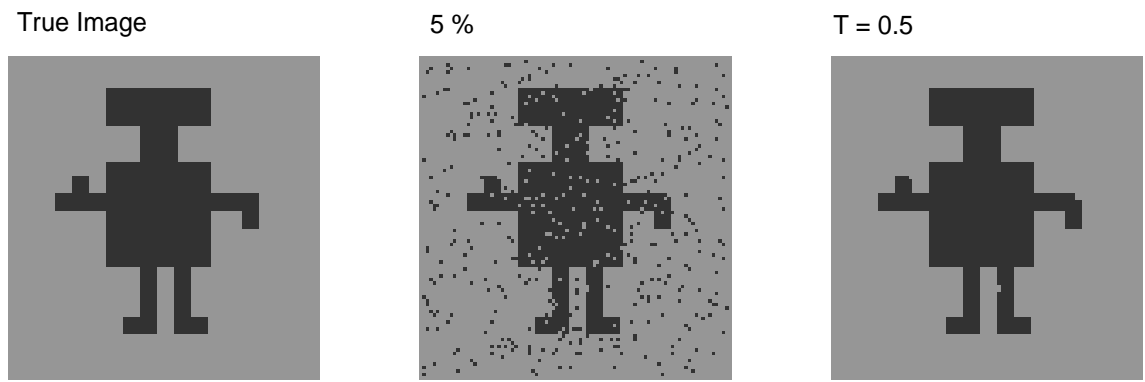}}
\caption{\small {\bf Binary ROBOT image; $L=92$. 
(Left) true image  $\hat{\Theta}$
(Middle) Image $X$ constructed by adding  noise to $\widehat{\Theta}$, 
as described in the section \ \lq Binary Symmetric Channel Likelihood
distribution\rq with $\tilde{p}=0.05$
and
(Right) Restored image, $\Theta_{{\rm MAP}}$. Complete 
restoration happens after one iteration.  
Image restoration  has been carried out at  $T=0.5$}}
\end{figure}
MAP estimates of the 5-gray level Potts image 
made at a high temperature  
($T=2.5$) and at a low temperature $0.01$ and at intermediate 
temperature ($T=0.51$) are shown in 
Fig. (8); the high temperature $\Theta_{{\rm MAP}}$ is
noisy and the low temperature $\Theta_{{\rm MAP}}$ has 
several of its 
fine features distorted. Good image restoration obtains 
for temperatures in the neighbourhood of $0.5$. 
\section*{Restoration of Benchmark Images}
\rhead[Restoration of  Toy and Benchmark Images]{\thepage}
\rhead[Markov Chain Monte Carlo Image Restoration]{\thepage}
\subsection*{Binary Lena Image}
We have taken a benchmark image called the Lena, painted with 
256 gray levels. We have converted it to a binary image  
and is displayed in 
Fig. (9 : Left). We introduce $5\%$ noise and the resulting 
image $X$ is also shown in Fig. (9 : Middle). 
We have employed Ising {\it Prior}
and Hamming distance and made an MAP estimate of the original image.
The restored image $\Theta_{{\rm MAP}}$
is depicted in Fig. (9 Right).

We have processed the same image at temperatures 
$T=0.1,\ 1.1,\ 2.5$ and the 
results are displayed in Fig. (10). 
At $T=0.1$, the algorithm has removed all the fine features 
of the image. At $T=2.5$ the algorithm has interpreted even noise as 
features and retained them. Good image restoration seems to happen at
temperatures between $1$ and $1.5$

\subsection*{Five and Ten Gray Level Lena Images}
We have coarse grained the same Lena image to $5$ 
gray levels and the result is shown in Fig. (11 : Left). 
The image corrupted with 
$5\%$ noise is depicted in Fig. (11 : Middle) 
and an MAP estimate made with 
Potts {\it Prior} and Hamming distance is depicted 
in Fig. (11 : Right).

Results of image restoration of Lena image with $10$ gray levels 
are shown in Fig. (12). 
It is clear from the discussions above such a search 
algorithm would be time consuming.
It would be advantageous to obtain an ensemble of 
images which has the desired {\it Posterior} 
distribution so that the required statistics 
can be calculated by simple 
arithmetic averaging over the {\it Posterior} ensemble.\\ 
\begin{figure}[htp]
\centerline{\includegraphics[height=2.0in,width=6.0in]{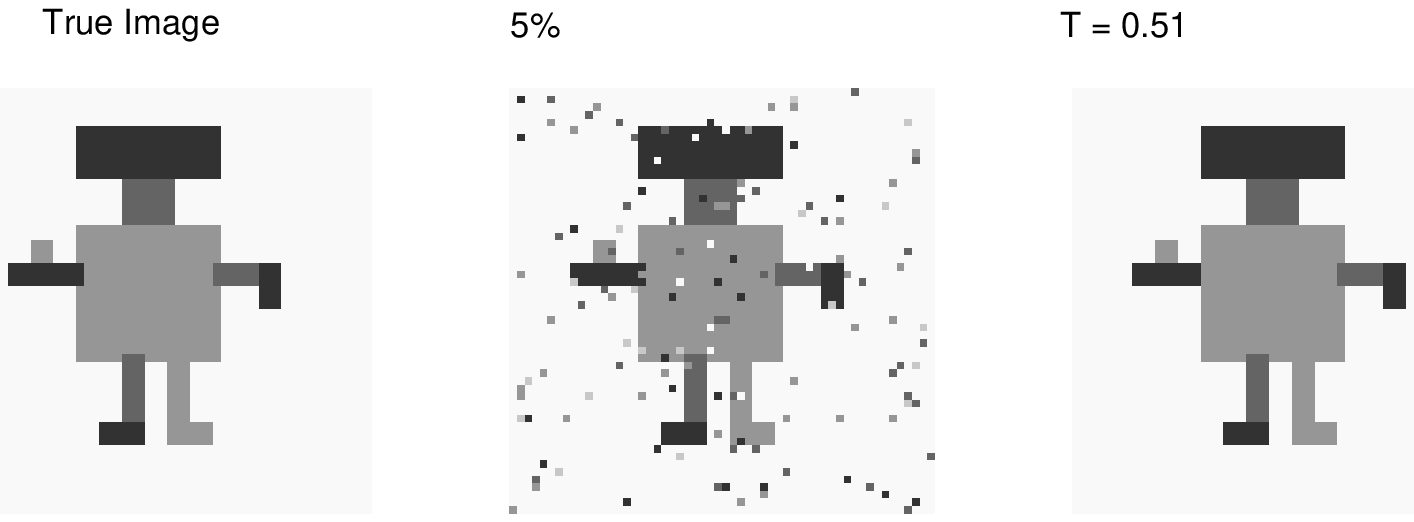}}
\caption{\small {\bf ROBOT image: L=56; $5$ gray levels; 
Image restoration has been carried out 
employing Potts {\it Prior}, 
Hamming {\it Likelihood}. 
(Left)  True image $\widehat{\Theta}$ 
(middle) Image  $X$ constructed by adding  noise to $\widehat{\Theta}$, 
as described in the section \ \lq Degradation of a multi-gray level image\rq\  
with $\tilde{p}=.05$, and 
(right)  restored image $\Theta_{{\rm MAP}}$ at $T=0.51$.} }
\vskip 20mm
\centerline{\includegraphics[height=4.5in,width=6.0in]
{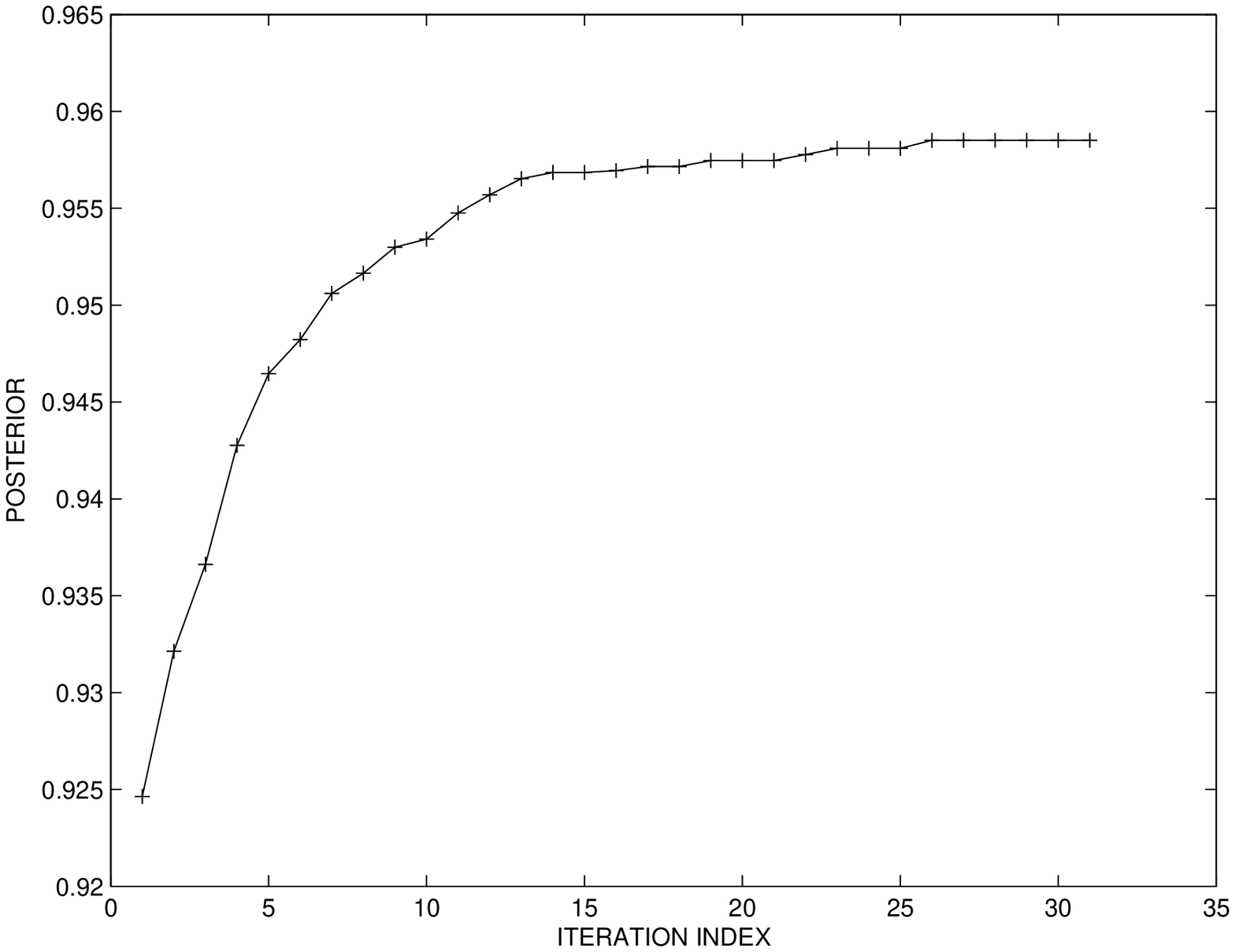}}
\caption{\small {\bf {\it Posterior}
maximum {\it versus} iteration index, for image restoration 
shown in Fig. (6)}}
\end{figure}
\newpage
\begin{figure}[htp]
\centerline{\includegraphics[height=2.0in,width=6.0in]{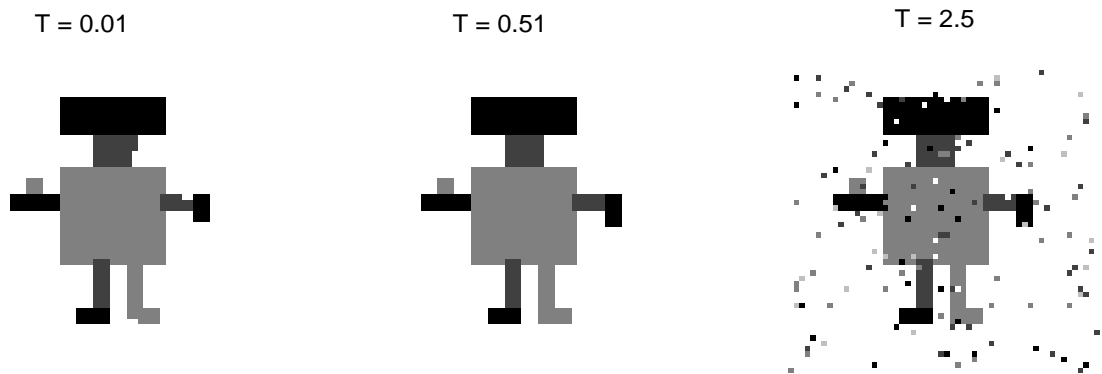}}
\caption{\small {\bf Robot image:  $L=56$;  $5$ gray levels.
Image restoration employs Potts {\it Prior}  
and Hamming {\it Likelihood}. The 
Maximum $\grave{ {\rm {\bf a}}}$ Posteriori (MAP) estimates at 
low : $T=.01$ (Left), intermediate  $T=0.51$  (middle) and 
high  : $T=2.5$ (right) temperatures are depicted. }
}\end{figure}
\begin{figure}[htp]
\centerline{\includegraphics[height=2.0in,width=6.0in]{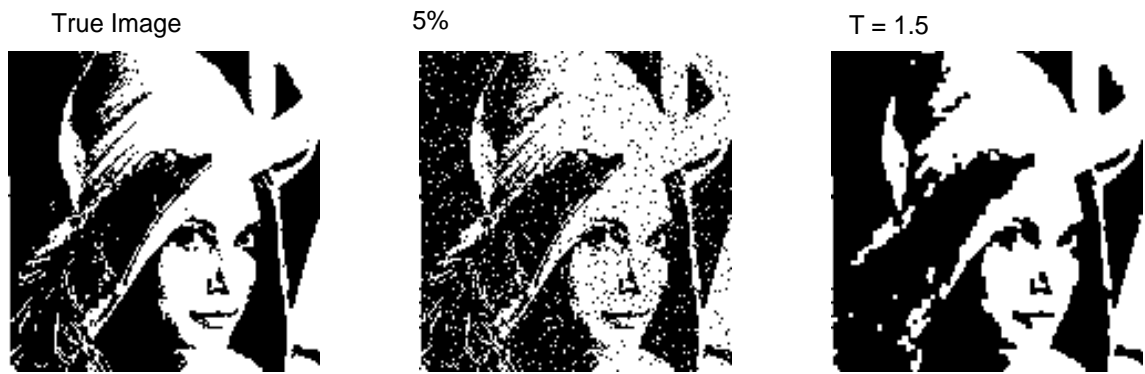}}
\caption{\small {\bf Restoration of the benchmark binary Lena image, 
employing Ising {\it Prior} and Hamming {\it Posterior}. 
$L=141$ 
(Left)   True  image  $\hat{\Theta}$  
(Middle) Image $X$ constructed by adding noise to  $\widehat{\Theta}$
         as described in section \ \lq Binary Symmetric Channel Likelihood 
distribution\rq\  with $\tilde{p}=.05$. 
(Right)   Restored image, $\Theta_{{\rm MAP}}$  after one  iteration.
image processing has been carried out at temperature $T=1.5$}}
\end{figure}
\newpage
\begin{figure}[htp]
\centerline{\includegraphics[height=2.0in,width=6.0in]{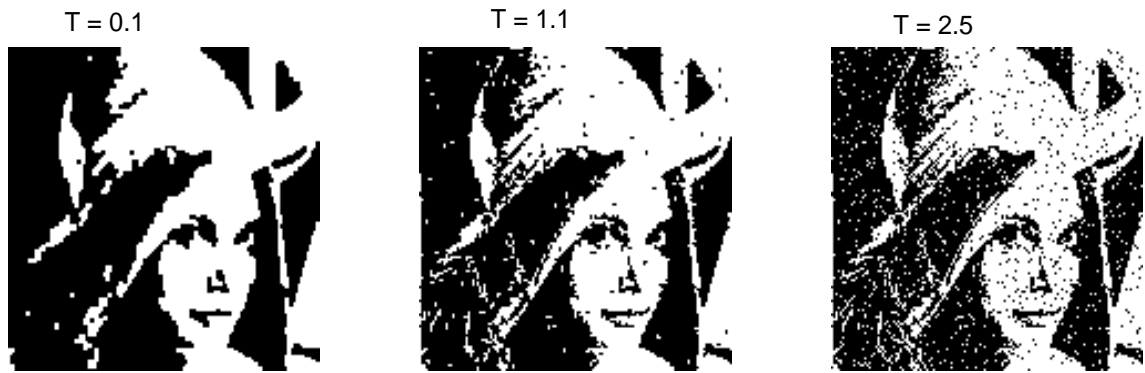}}
\caption{\small {\bf Restoration of binary Lena image at 
three temperatures. MAP estimates of $\widehat{\Theta}$ are given. 
(Left) $T=0.1$ (Middle) $T=1.1$ and 
(Right) $T=2.5$. $\widehat{\Theta}$ and $X$ are the same ones depicted in 
Fig. (9). }}
\end{figure}
\begin{figure}[htp]
\centerline{\includegraphics[height=2.0in,width=6.0in]{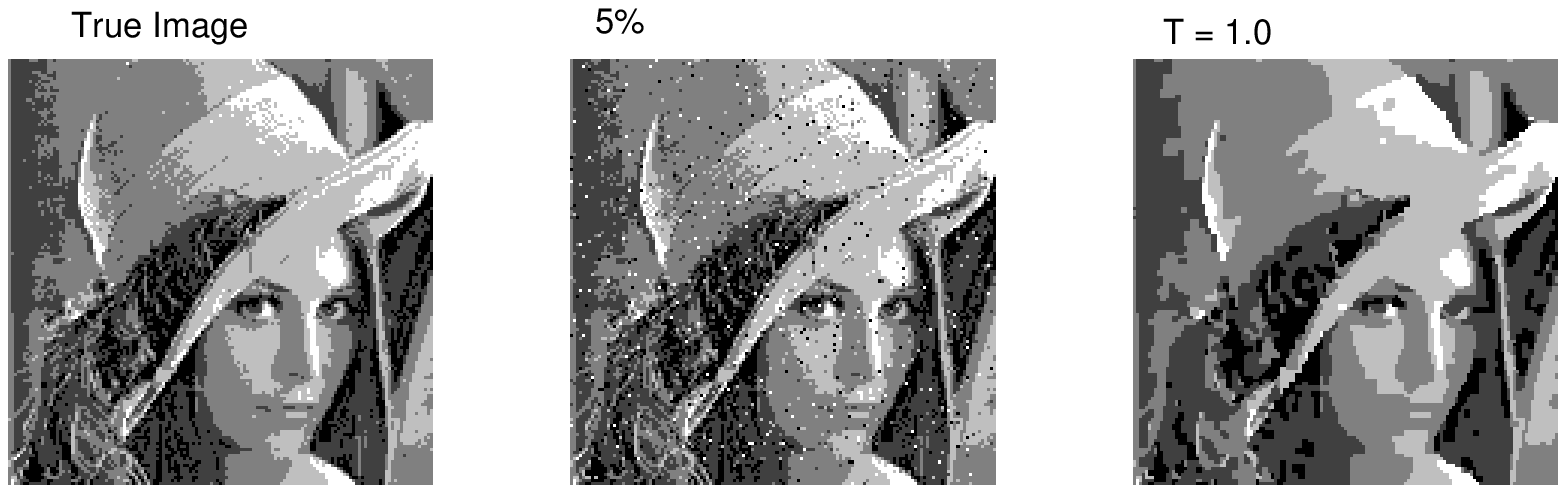}}
\caption{\small {\bf Restoration of the benchmark 5-gray level
Lena image employing Potts {\it Prior} and Hamming distance. 
(Left) True image $\widehat{\Theta}$ (Middle) Image $X$ constructed by adding noise to 
$\widehat{\Theta}$ as described in section\  \lq Degradation of a multi-gray level image\rq\ 
with $\tilde{p}=0.05$. 
The image restoration was done at T=1.0}}
\end{figure}
\begin{figure}[htp]
\centerline{\includegraphics[height=2.0in,width=6.0in]{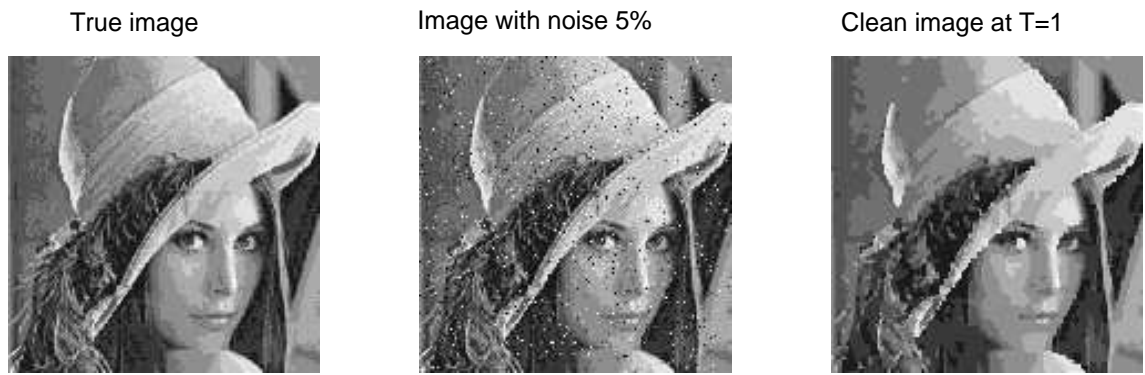}}
\caption{\small {\bf Restoration of the benchmark 10-gray level
Lena image employing Potts {\it Prior} and Hamming distance. 
(Left) True image $\widehat{\Theta}$ (Middle) Image $X$ constructed by adding noise to 
$\widehat{\Theta}$ as described in section \ \lq Degradation of a multi-gray level image\rq\ 
with $\tilde{p}=0.05$.
The image restoration was done at T=1.0}}
\end{figure}
It is precisely in this context that the 
Metropolis algorithm\cite{Metropolis}, discovered 
for purpose of obtaining a  canonical ensemble of 
microstates in statistical mechanics becomes useful. 
We outline below a Markov chain Monte Carlo 
technique in conjunction with the Metropolis 
algorithm and its variant, to obtain a
{\it Posterior} ensemble of images.
\section*{Markov Chain Monte Carlo for Sampling from $\grave{{\rm{\bf a}}}$
Posteriori Distribution}
\rhead[Markov Chain Monte Carlo Image Restoration]{\thepage}
Monte Carlo is a numerical technique that makes use 
of random numbers to solve a problem.
Historically, the first large scale Monte 
Carlo work carried out dates back to the middle of twentieth century.
This work pertained to simulation of neutron multiplication,
scattering, streaming and eventual absorption in a medium or escape from
it. Application of Monte Carlo
method to  problems in statistical mechanics 
started with the discovery of the 
Metropolis algorithm \cite{Metropolis}
which generates a Markov chain of microstates 
converging to the desired ensemble. There exist a vast
body of literature on Monte Carlo  technique. 
We refer to \cite{JMHDCH,DPLKB,IMS,
FJ,KBDWH,PC,KPNM_2000,KPNM_2004} for some of them.
Monte Carlo constitutes a   
natural numerical technique for image processing. We describe below 
a Markov Chain Monte Carlo algorithm for image restoration.

We fix the temperature $T$. 
We start with an arbitrary image $\Theta_0$. 
A good choice of $\Theta_0$ is $X$.
Then we construct a Markov chain  
whose asymptotic segment contains images 
belonging to the desired {\it Posterior} ensemble at 
the chosen temperature. In a Markov chain
$\Theta_0\to\Theta_1\to\cdots\to\Theta_n\to\cdots$, 
the image $\Theta_{k+1}$ depends 
only on $\Theta_k$ and not on the previous history.  
In  Appendix 2
we have described briefly, the basic elements of a 
Markov chain and its construction by 
Monte Carlo algorithms. For a detailed description  of Markov chain
see \cite{Noris}. Here we describe only the operational 
details  of a Markov chain Monte Carlo
algorithm.

First calculate the {\it Posterior}, $\pi(\Theta_k\vert X)$ (upto a normalization constant) of the 
current  image $\Theta_k$; denote it by the symbol $\pi_k$. 
Select randomly a pixel from the image plane.  Change its  gray level 
randomly\footnote{For example if we are processing a binary image, then switch the gray level of the 
chosen pixel from its present value to the other; if we are processing an
image with $Q$ gray levels, then change the gray level of the chosen pixel to one of the 
$Q$ values randomly.}, and get a trial image $\Theta_t$. 
Calculate
the {\it Posterior} of the trial image, $\pi(\Theta_t\vert X)$ and denote it by $\pi_t$.
Accept the trial image with a probability ${\hat p}$ given by,
\begin{eqnarray}
\hat{p}={\rm min}\Big(1,\ \frac{\pi_t}{\pi_k}\Big)
\end{eqnarray}
This is called the Metropolis algorithm \cite{Metropolis}. This is also known as 
Metropolis-Hastings algorithm in image processing literature\footnote{Metropolis algorithm is 
a special case of a more general Hastings algorithm \cite{WKH}.}. 
We notice that the Metropolis acceptance
is based on the ratios of the {\it Posteriors}.
The normalization constants cancel.  Hence
it is adequate if we know the {\it Posterior} upto a normalization constant.
 
We can also employ  
Gibbs' sampler\footnote{The Gibbs sampler was discovered in statistical physics by Cruetz \cite{MC} and
is known by the name heat-bath algorithm; it has also been discovered independently 
in the context of spatial statistics by Ripley \cite{BDR}, Grenader \cite{UG} and Gemen and Gemen \cite{SGDG}.
The name Gibbs sampler is due to Gemen and Gemen \cite{SGDG}.
The Glauber algorithm \cite{Glauber} is the same as heat-bath algorithm, but for a minor 
irrelevant detail \cite{KPNM_2004}.}
in which the acceptance probability is given by
\begin{eqnarray}
\hat{p}=\frac{\pi_t}{\pi_k + \pi_t}.
\end{eqnarray}
The acceptance/rejection step  
is implemented as follows.  We call a random number $\xi$, distributed 
uniformly in the range zero to unity. If $\xi \le {\hat p}$, 
accept the trial image and set $\Theta_{k+1} =\Theta_t$. Otherwise
reject the trial image and set $\Theta_{k+1}=\Theta_k$. 
Repeat the above on the image $\Theta_{k+1}$. Iterate and construct a Markov chain of images
given by
\rhead[Markov Chain Monte Carlo]{\thepage}
\rhead[Markov Chain Monte Carlo Image Restoration]{\thepage}
\begin{eqnarray}
\Theta_0\to\Theta_1\to\Theta_2\to\cdots\to\Theta_n\to\Theta_{n+1}\to\cdots
\end{eqnarray}
In practice, we define a consecutive set of $N$ attempted updates (successful or otherwise) as constituting 
a Monte Carlo Sweep (MCS), where  $N$ is the total number of pixels in the image being processed.
We take the image  at the end of successive MCS 
and construct a Markov chain.  We have shown in  Appendix 2 that the asymptotic 
part of the chain $\{ \Theta_m\ :\ m\ge n\to\infty\}$ contains images that belong the desired {\it Posterior} ensemble.
Let $\Gamma$ denote the set of images taken from the asymptotic segment of the Markov chain.
$\Gamma$ is called the {\it Posterior}  ensemble. Let $\widehat{\Gamma}$ denote the total number of images 
in $\Gamma$. MAP, MPM and TPM estimates of $\widehat{\Theta}$
can be made from the {\it Posterior} ensemble  $\Gamma$ as follows.
\begin{eqnarray}
\Theta_{{\rm MAP}}={\rm arg}\ \begin{subarray}
                                 \ \cr
                                   \ \cr
                                      {\rm max}\cr
                                      \Theta\in\Gamma
                                      \end{subarray}\ \ 
                                      \pi(\Theta\vert X)
\end{eqnarray}
For calculating MPM estimate, we partition $\Gamma$ into mutually exclusive and exhaustive subsets of images
as described below. Consider a pixel $i\in{\cal S}$. Define $\Gamma_{\zeta}^{(i)}$ as a subset of images
for which the gray level of pixel $i$ is $\zeta$, where $\zeta=0,1\cdots Q-1$:
\begin{eqnarray}
\Gamma_{\zeta}^{(i)}=\Big\{ \Theta \in \Gamma, \theta_i(\Theta)=\zeta\Big\}\ {\rm for}\ \zeta=0,1,\cdots,Q-1
\end{eqnarray}
Let $\widehat{\Gamma}_{\zeta}^{(i)}$ denote the number of images belonging to the 
subset $\Gamma_{\zeta}^{(i)}$. Then we have,
\begin{eqnarray}
\zeta_{{\rm MPM}}^{(i)}&=&{\rm arg}\ \begin{subarray}
                                  \ \cr
                                   \ \cr
                                    {\rm max}\cr
                                    \ \ \zeta
                                \end{subarray}
                              \ \ \widehat{\Gamma}_{\zeta}^{(i)}\\
   &  & \nonumber\\
\zeta_{{\rm TPM}}^{(i)}&=&{\rm arg}\ \begin{subarray}
                                   \ \cr
                                   \ \cr
                                  {\rm min}\cr
                                   \ \zeta
                                   \end{subarray}
                                   \ \bigg[ \zeta-\frac{1}{\widehat{\Gamma}}
                                     \sum_{\Theta\in\Gamma}\theta_i(\Theta)\bigg]^2
\end{eqnarray}
In other words, $\zeta_{{\rm MPM}}^{(i)}$ is the value of $\zeta$ for which 
$\widehat{\Gamma}_{\zeta}^{(i)}$ is maximum, while $\zeta_{{\rm TPM}}^{(i)}$ is the 
value of $\zeta$ closest to $\overline{\theta}_i$ - the arithmetic average
of the gray level of pixel $i$ taken over images belonging to
the {\it Posterior}
ensemble $\Gamma$. 
We depict in Fig. (13) MAP, MPM and TPM estimates for a binary ROBOT image
with  $L=56$.
\begin{figure}[htp]
\centerline{\includegraphics[height=2.in,width=6.0in]{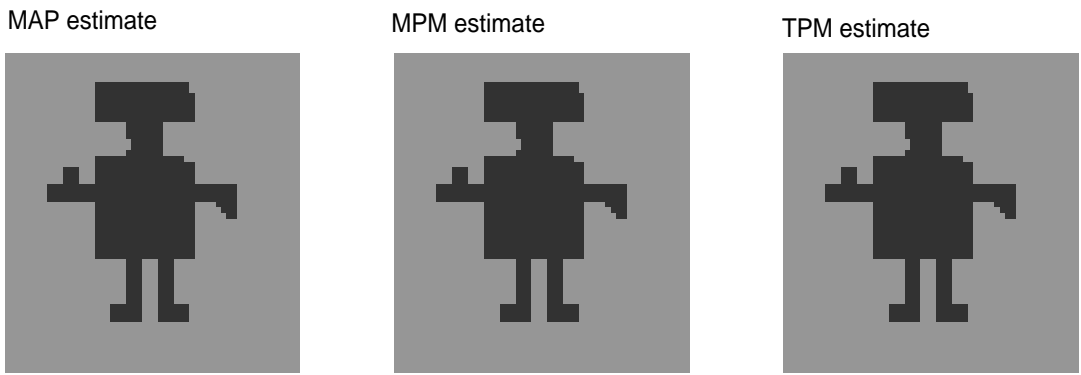}}
\caption{\small {\bf Image restoration employing Ising {\it Prior}, Hamming {\it Likelihood} and Markov
Chain Monte Carlo with Metropolis acceptance. 
MAP (Left), MPM (middle) and  TPM (Right) estimates  
made at $T=0.5$ are shown. These statistics have been obtained from the ensemble of images sampled 
employing Markov chain Monte Carlo in conjunction with the Metropolis algorithm. 
The $\Theta_{{\rm{\bf MAP}}}$ obtained from an earlier Monte Carlo search algorithm 
was taken as input to avoid removing initial images from the Markov chain for 
purpose of equilibration. All the three estimates give identical results. In any case
for binary image one can show that by definition MPM and TPM estimates are the same} }
\end{figure}

We have  not presented in this review detailed results on the MPM and TPM estimates
of the true image, because we find that these estimates are not good if we have only 
a single hyper-parameter in the image restoration algorithm 
We find that retaining two independent hyper-parameters one from the 
{\it Likelihood} model of degradation ($\beta_L$) and the other from the {\it Prior}
model of out subjective expectation ($\beta_P$) would be required for a meaningful
estimate of $\widehat{\Theta}$ through MPM and TPM statistics employing Markov chain 
Monte Carlo technique in conjunction with Metropolis algorithm or Gibbs sampler.
We need to consider  images that are consistent and that are not consistent with the 
{\it Likelihood} and {\it Prior} model assumptions and investigate the 
performance of Markov chain Monte Carlo restoration
algorithms. Work in this direction is in progress and would be communicated soon.

In the  Metropolis or the Heat-bath algorithms described above, 
only a single pixel is updated at a time. For restoring a large image,  
single pixel update algorithms can be frustratingly  time consuming. Also 
if there are strong correlations in an image, 
which often is the case, we could  think of updating the states (gray levels)  
of a large cluster of pixels, to speed up the algorithm. If we take an arbitrary 
cluster of pixels of the same gray level and update their gray levels coherently, most often the trial image 
constructed would get rejected in the Metropolis step. Hence it important to have a 
correct definition of a cluster.  
It is in this context a study of  the cluster algorithm proposed by 
Swendsen and Wang \cite{SW} 
becomes relevant.

\section*{Swendsen-Wang Cluster Algorithm}
\rhead[Cluster Algorithms]{\thepage}
\rhead[Markov Chain Monte Carlo Image Restoration]{\thepage}
Swendsen and Wang \cite{SW} derived an ingenious cluster algorithm for mapping
the Ising/Potts spin problem to a bond-percolation problem based on the
work of Kasteleyn and Fortuin \cite{PWKCMF} and Coniglio and Klein \cite{ACWK}. 
The algorithm was originally intended for overcoming the problem of 
critical slowing down near second-order phase transition. This algorithm has been 
adapted to several problems in image processing since recent times. 
A comparison of Swendsen-Wang algorithm and Gibbs sampler  
was reported in \cite{AJG}. The dynamics of 
Swendsen-Wang algorithm in image segmentation has been investigated in \cite{IG}. 
Swendsen-Wang algorithm has found applications  in Bayesian variable selection
particularly for problems where there are multi-colinearities amongst predictors 
\cite{DJNPJG}. Very recently a version of Swendsen-Wang algorithm for removing 
Poisson noise from medical images was proposed \cite{SLLWN}.  In fact it has been 
recognized  that the Swendsen-Wang algorithm is a particular example
of a more general auxiliary variable method pioneered by 
Edwards and Sokal \cite{RGEADS} and Besag and Green \cite{JBPJG}. For a review 
of auxiliary variable method see \cite{DMH}.
\subsection*{Auxiliary Bond Variables}
Let us first consider sampling of images from the 
Ising/Potts {\it Prior}. Introduce a collection ${\mathbf \epsilon}$, 
of binary random variables : ${\mathbf \epsilon}=\{ \epsilon_{i,j}:i,j\in{\cal S}\}$
where $i,j$ are nearest neighbour pixels. These
are called auxiliary bond random variables. 
$\epsilon_{i,j}=0\ \  {\rm or}\ \ 1$. If
$\theta_i=\theta_j$ then we say there exists a {\it satisfied} bond between $i$ and $j$.
A {\it satisfied} bond may be {\it occupied} ($\epsilon_{i,j}=1$) or 
may {\it not be occupied} ($\epsilon_{i,j}=0$).
If $\theta_i\ne\theta_j$ then there is 
no bond connecting the pixels $i$ and $j$. In other words
$\epsilon_{i,j}=0$, if $\theta_i\ne\theta_j$.
Let $p$ denote the probability of occupying a satisfied bond  in  $\Theta$.  The
probability of not occupying a satisfied bond is $q=1-p$, so that $p+q=1$.
We do not yet know the value of $p$ we must use for obtaining an ensemble of 
images distributed as per the Ising/Potts {\it Prior}. We shall take up this question
after going through the following preliminaries. 
\subsubsection*{Conditional, Joint and Marginal Priors}
Construction of  a bond structure on the given image, described above,
can be expressed  mathematically as,
\begin{eqnarray}
\pi(\epsilon_{i,j}=1\vert\theta_i=\theta_j)&=& p,\nonumber\\
\ \nonumber\\
\pi(\epsilon_{i,j}=0\vert\theta_i=\theta_j)&=& q,\nonumber\\
\ \nonumber\\
\pi(\epsilon_{i,j}=1\vert\theta_i\ne\theta_j)&=& 0,\nonumber\\
\ \nonumber\\
\pi(\epsilon_{i,j}=0\vert\theta_i\ne\theta_j)&=& 1.
\end{eqnarray}
The conditional {\it Prior} in terms of $p$ and $q$ is given by,
\begin{eqnarray}\label{conditional_Prior_epsilon_given_theta}
\pi(\epsilon\vert\Theta)=\prod_{\langle i,j\rangle}\bigg[ \bigg\{ 
q\ {\cal I}(\epsilon_{i,j}=0)+
p\  {\cal I}(\epsilon_{i,j}=1)\bigg\} {\cal I}(\theta_i=\theta_j)
+{\cal I}(\epsilon_{i,j}=0)\ {\cal I}(\theta _i \ne\theta _j)\bigg].
\end{eqnarray}
The joint {\it Prior} is given by,
\begin{eqnarray}\label{joint_prior}
\pi(\Theta ,{\mathbf\epsilon}) = 
\frac{1}{Z_1}\prod_{\langle i,j\rangle}\bigg[ q\  {\cal I}
(\epsilon_{i,j}=0)+p\ {\cal I}(\epsilon_{i,j}=1)\ {\cal I}(\theta_i=
\theta_j)\bigg].
\end{eqnarray}
where $Z_1$ is the normalization constant.
From the joint {\it Prior} $\pi (\Theta,\epsilon)$, we can derive an expression for the marginal {\it Prior} 
$\pi(\theta)$ 
and is given by
\begin{eqnarray}\label{marginal_Prior_theta}
\pi(\Theta)&=& \sum_{\epsilon}\pi(\Theta,\epsilon)\nonumber\\
\ \nonumber\\
           &=&\frac{1}{Z_1}\prod_{\langle i,j\rangle}\Big[ p{\cal I}(\theta_i=\theta_j)\ +\ q\Big]
\end{eqnarray}
\noindent
{\bf What is the  value of $p$ appropriate  for sampling from the Ising/Potts Prior ?}
\vskip 3mm
\noindent
Let us start with the Ising/Potts {\it Prior} given by Eq. (\ref{Ising-Potts_Prior}) and 
express it in a different form
as described below.
\begin{eqnarray}\label{prior_2}
\pi (\Theta)&=&\frac{1}{Z(\beta)}\  
\exp\Big[ -\beta\sum_{\langle i,j\rangle}{\cal I}(\theta_i\ne\theta_j)\Big]
\nonumber\\
\ \nonumber\\
\ \nonumber\\
&=&\frac{1}{Z(\beta)}\ \prod_{\langle i,j\rangle}
               \Big[ {\cal I}(\theta_i=\theta_j)+e^{-\beta}
               {\cal I}( \theta_i\ne\theta_j)\Big]\nonumber\\
\ \nonumber\\
\ \nonumber\\
&=&\frac{1}{Z(\beta)}\ \prod_{\langle i,j\rangle}\Big[ {\cal I}(\theta_i=\theta_j)+
\Big\{ 1-{\cal I}(\theta_i=\theta_j)\Big\}\times e^{-\beta}\Big]\nonumber\\
\ \nonumber\\
\ \nonumber\\
&=&\frac{1}{Z(\beta)}\ \prod_{\langle i,j\rangle}\Big[ \big( 1-e^{-\beta}\big) {\cal I}(\theta_i=\theta_j)
+\exp(-\beta)\Big]
\end{eqnarray}
Comparing  the above with  the marginal {\it Prior} $\pi(\Theta\vert\epsilon)$ given by
Eq. (\ref{marginal_Prior_theta}) we can conclude that
$p=1-\exp(-\beta)$, $q=\exp(-\beta)$  and the Ising/Potts partition function $Z(\beta)$ is the same as
the normalization constant  of the joint 
{\it Prior} $Z_1$\footnote{For the Ising model in statistical mechanics, 
$p=1-\exp(-2\beta)$, since the 
ground state and the excited state of a pair of nearest neighbour Ising spins differ by
two units of energy, {\it i.e.} $\Delta E=2$. However, for Potts spin model $p=1-\exp(-\beta)$, 
since $\Delta E=1$. In both the Ising and Potts {\it Prior} models considered here 
$\Delta E=1$ and hence $p=1-\exp(-\beta)$.}.

Another way  of looking at the same thing, which gives  a better insight 
into the Swendsen-Wang algorithm, 
is as follows.
Let $N_E$ denote the total number of distinct nearest neighbour pairs of pixels in  the image plane.
If a pixel is considered as a vertex, then $N_E$ is the number of edges in the graph.
Let $B$ denote the total number of like pairs of pixels in an image $\Theta$; {\it i.e.} $B$ is the total number of
{\it satisfied} bonds in  $\Theta$. The Ising/Potts {\it Prior} can be written in terms of $N_E$ and $B$
as follows. Consider Eq. (\ref{Ising-Potts_Prior}). The summation in the exponent of the Ising/Potts {\it Prior} can be split
into a sum over 
like pairs and a sum over unlike pairs,
\begin{eqnarray}
\sum_{\langle i,j\rangle}{\cal I}(\theta_i\ne\theta_j)&=&\sum_{\begin{subarray}
                                                            \ \langle i,j\rangle\cr
                                                              \theta_i=\theta_j
                                                                \end{subarray}}
                  {\cal I}(\theta_i\ne\theta_j)
                  + \sum_{\begin{subarray}
                                                            \ \langle i,j\rangle\cr
                                                              \theta_i\ne\theta_j
                                                                \end{subarray}}
                  {\cal I}(\theta_i\ne\theta_j)\nonumber\\
\ \nonumber\\
                  &=& N_E -B
\end{eqnarray}
Thus we get,
\begin{eqnarray}\label{prior_1}
\pi (\Theta)&=&\frac{1}{Z(\beta)}\exp\big[ -\beta(N_E -B)\big]
\end{eqnarray}
Let $b$ denote the number of occupied bonds in the image $\Theta$. Note that $b$ is a random variable. 
For an image $\Theta$, the value of  $b$ can range from $0$ to $B$. It is easily seen that the joint {\it Prior}
$\pi(\Theta,\epsilon)$ can be expressed in terms of $N_E$ and $b$ as,
\begin{eqnarray}
\pi(\Theta ,{\mathbf\epsilon})=\frac{1}{Z_1}q^{N_E -b}p^b
\end{eqnarray}
Integrate $\pi(\Theta,{\bf \epsilon})$ over ${\bf\epsilon}$ (keeping in mind that 
we need to sum over the distribution of $b$)  and get
the  marginal distribution of $\Theta$ in terms of $N_E$ and $B$:
\begin{eqnarray}\label{prior_3}
\pi(\Theta)&=&\frac{1}{Z_1}\ \sum_{b=0}^{B}\ 
\frac{B!}{b!\  (B-b)!}\ q^{N_E-b}\ p^b\nonumber\\
\ & & \nonumber\\
\ & & \nonumber\\
           &=&\frac{1}{Z_1}\  q^{N_E-B}\ \sum_{b=0}^{B}\ 
\frac{B!}{b!\  (B-b)!}\ q^{B - b}\ p^b\nonumber\\
\ & & \nonumber\\           
\ & & \nonumber\\
&=&\frac{1}{Z_1}\ q^{N_E-B}
\end{eqnarray}
The above is the same as the Ising/Potts {\it Prior} given by Eq. (\ref{prior_1}), if we identify that
$Z_1=Z(\beta)$, $q=\exp(-\beta)$ and hence $p=1-\exp(-\beta)$.
We also could have come to the same conclusion directly from the 
marginal {\it Prior} given by Eq. (\ref{marginal_Prior_theta}),
\begin{eqnarray}
\pi(\Theta)=\frac{1}{Z_1}\ (p+q)^B\ q^{N_E-B}\ =\ \frac{1}{Z_1}\ q^{N_E-B}.
\end{eqnarray}

\subsection*{Conditional $\grave{{\rm{\bf a}}}$  Priori Distribution
 $\pi(\Theta\vert\epsilon)$}
For practical implementation of the algorithm, we need  conditional
{\it Priors}: $\pi(\epsilon\vert\Theta)$ and $\pi(\Theta\vert \epsilon)$. 
The simulation strategy is to sample alternately from these two conditional {\it Priors} and construct a 
Markov chain of images which asymptotically converges to the {\it Prior} ensemble.
We already have an expression for $\pi(\epsilon\vert\Theta)$
given by Eq. (\ref{conditional_Prior_epsilon_given_theta}), with
$q=\exp(-\beta)$ and $p=1-\exp(-\beta)$. 

To derive an expression for the conditional {\it Prior} $\pi(\Theta\vert\epsilon)$, we proceed as follows.
We have,
\begin{eqnarray}
\pi(\Theta\vert \epsilon)&=&\frac{\pi(\Theta,\epsilon)}{\pi(\epsilon)}\nonumber\\
                         & & \nonumber\\
\ & & \nonumber\\
&=&\frac
{
Z_1^{-1}\prod_{\langle i,j\rangle}\Big[ q\ {\cal I}(\epsilon_{i,j}=0)+p\ {\cal I}(\epsilon_{i,j}=1)
{\cal I}(\theta_i=\theta_j)\Big]
}
{
Z_1^{-1}\Big( \prod_{
\begin{subarray}
\ \ \langle i,j\rangle\cr
\epsilon_{i,j}=0
\end{subarray}}\ \ q\Big)
\Big( \prod_{\begin{subarray}
\ \ \langle i,j\rangle\cr
\epsilon_{i,j}=1\end{subarray}}\ \ p\Big) Q^{N_C (\epsilon)}
}\nonumber\\
& & \nonumber\\
\ \nonumber\\
  &=&\frac{1}{ Q^{N_C(\epsilon)}}
\end{eqnarray}
where $N_C$ is the number of clusters generated by imposing $\epsilon$ bonds on the image.
Essentially we have converted an interacting system  into a non interacting cluster system. Each cluster
acts independently. We switch the gray level of each cluster of pixels randomly, coherently and independently. 
 
\subsection*{Sampling $\epsilon$  from $\pi(\epsilon\vert\Theta_k)$}
Let $\Theta_k$ denote the current image in the Markov chain.
Take a pixel $i$; this constitutes a single-pixel cluster. Let $\nu_i$ denote the 
set of pixels that are nearest neighbours of $i$. Take a pixel $j\in\nu_i$. 
If $\theta_j(\Theta_k)\ne\theta_i(\Theta_k)$ do not put a bond between them. On the 
other hand if $\theta_j(\Theta_k)=\theta_i(\Theta_k)$ we say there exists a satisfied 
bond between the two pixels. Then call a random number 
$\xi$. If $\xi \le p=\exp(-\beta)$ then put a bond between them. In other words we occupy the 
satisfied bond with a probability $p$. If the decision is to occupy the 
satisfied bond then add the pixel to the cluster and form a two-pixel cluster.
Repeat the above on all the nearest neighbours of $i$ and on all the nearest neighbours 
of pixels that get added to the cluster. The cluster growth 
will eventually terminate.
Thus we get a cluster grown from the seed pixel $i$ and let us denote 
the cluster by the symbol $c_1$. . 
Start with a new pixel that does not belong to $c_1$. Grow a cluster in the region of the
image plane excluding the one  occupied by $c_1$. Call this cluster $c_2$. 
The process of growing cluster from new seeds would eventually stop when all the 
pixels in the image plane have been assigned with one cluster label or other. 
Thus we have non-overlapping and exhaustive set of clusters on the 
image plane.
 
\subsection*{Sampling $\Theta$ from $\pi(\Theta\vert\epsilon)$}
Change the gray levels of all pixels in a cluster to a gray level
randomly chosen from amongst the $Q$ gray levels. Carry out this process on each cluster
independently. Remove the bonds in the resulting image  and call it  $\Theta_{k+1}$.

\subsection*{Construction of Prior Ensemble}
Start with $\Theta_0=X$. Form clusters on the image plane, 
by sampling $\epsilon$  from $\pi(\epsilon\vert\Theta_k)$
as described above. Get a new image $\Theta_1$ by sampling from $\pi(\Theta\vert\epsilon)$
as described above. Iterate and get a Markov chain of images. The asymptotic 
part of the Markov chain would contain images belonging to the {\it Prior}
ensemble. 

Next,  we turn our attention to adapting the Swendsen-Wang algorithm
to image restoration wherein  we need to sample images from the {\it Posterior} and not  
from the {\it Prior}. The key point is that having formed clusters on the image plane
of $\Theta_k$
we can get $\Theta_{k+1}$ by sampling  
 the cluster gray levels from the {\it Likelihood} distribution independently. The whole 
procedure is described below.

\section*{Posterior Ensemble: Swendsen-Wang  Algorithm}
Start with an arbitrary image $\Theta_0$. A good choice is $\Theta_0=X$. 
Sample bond variables $\epsilon$ from the conditional {\it Prior} $\pi(\epsilon\vert\Theta_0)$ 
as described earlier. We get clusters of pixels. 

Consider one of the 
clusters, say the $k$-th cluster. Let $\eta_k$ denote the pixels belonging to the 
cluster.  Sample a  gray level for this cluster from the {\it Likelihood} distribution,
\begin{eqnarray}
p_{\zeta}&=&{\cal L}\Big( \{ \theta_i=\zeta\ :\ i\in\eta_k\}
\Big\vert \{ x_i\ :\ i\in\eta_k\}\Big),\nonumber\\
& & \nonumber\\
&=&\frac{\exp\Big[ -\sum_{i\in\eta_k}f_i(\theta_i=\zeta,x_i)\Big] }
          {\sum_{\zeta=1}^{Q}\exp \Big[ -\sum_{i\in\eta_k}f_i (\theta_i=\zeta,x_i)\Big] }.
\end{eqnarray}
Sampling from the discrete distribution 
$\{ p_{\zeta}\ :\ \zeta=0,\ 1,\ \cdots ,\ Q-1 \}$ is carried out as follows.
Calculate the cumulative distribution
\begin{eqnarray}
P_0&=&0\nonumber\\
P_k&=&\sum_{m=1}^{k}p_m\nonumber\\
P_Q&=&1
\end{eqnarray}
Let $\xi$ be a random number uniformly distributed between $0$ and $1$. The gray level $k$ for which
$P_{k}\  <\ \xi\ \le\ P_{k+1}$ is assigned to  the 
pixels in the chosen cluster. 
Repeat the exercise  for all the clusters independently. Remove the 
bonds. The resulting image $\Theta_1$ is the image for the next graph construction. Iterating we get a 
Markov chain,
\begin{eqnarray}
&  &\Theta_0\ \to\ (\Theta_0,\epsilon_0)\ \to\ (\Theta_1,\epsilon_0)\ \to\ \Theta_1\ \to\ (\Theta_1,\epsilon_1)\ \to\ 
(\Theta_2,\epsilon_1)\ \to\ \Theta_2\cdots\ \to\ \nonumber\\
\ &\ &\nonumber\\
\to&  &\Theta_n\ \to\ (\Theta_n,\epsilon_n)\ \to\ (\Theta_{n+1},\epsilon_n)\ \to\ \Theta_{n+1}\ \to\ \
\cdots\nonumber 
\end{eqnarray}
which converges asymptotically to the to the {\it Posterior} ensemble.
We can make an MAP, an MPM or a TPM estimate the true image $\widehat{\Theta}$ from the 
{\it Posterior} ensemble.

\section*{Posterior Ensemble: Wolff's Algorithm}
Wolff \cite{Wolff} proposed a simple modification to the 
Swendsen-Wang cluster algorithm. Wolff's cluster algorithm can be 
adapted to image restoration as described below.

A single cluster is grown from a randomly chosen
pixel employing Swendsen-Wang prescription. The gray level
of all the pixels in that  cluster is updated 
by sampling from the {\it Likelihood} distribution. 
This results in a new  image that constitutes the 
next entry in the Markov chain. A pixel is chosen randomly in the 
new image and the whole process is repeated. A Markov chain of 
images is constructed. The asymptotic part of the Markov chain 
would contain images that belong to the desired {\it Posterior} ensemble.
An estimate of $\widehat{\Theta}$ can be made from the {\it Posterior} 
ensemble employing MAP or MPM or TPM statistics. 

We have employed Wolff's cluster algorithm in restoration of 
a binary and a $5$-gray level ROBOT image, at $T=0.51$. 
The results are depicted in Figures (14) and (15). 
\begin{figure}[ht]
\centerline{\includegraphics[height=2.0in,width=6.0in]{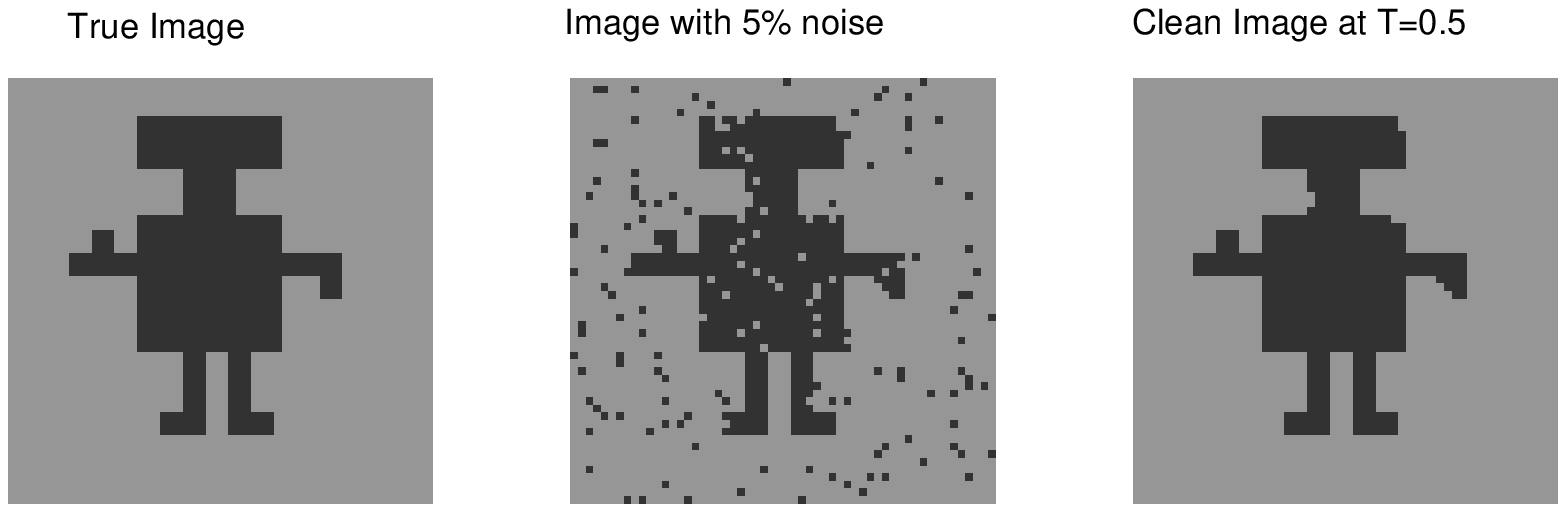}}
\caption{\small {\bf Image restoration employing 
Ising {\it Prior}, Hamming {\it Likelihood} and 
Wolff cluster algorithm
for estimating $\Theta_{{\rm MAP}}$, for a binary ROBOT image. 
(Left) $\widehat{\Theta}$ (middle) $X$ (Right) MAP estimate. 
image processing has been carried out at temperature $T=0.5$}}
\vskip 20mm
\centerline{\includegraphics[height=2.0in,width=6.0in]{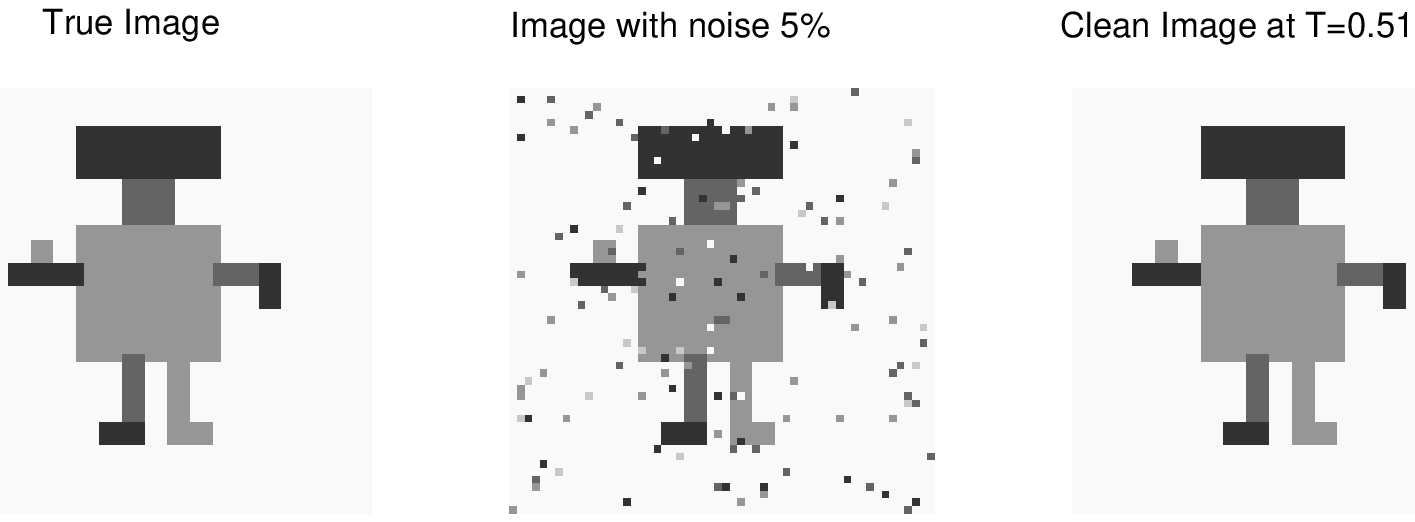}}
\caption{\small {\bf Image restoration employing 
Potts ($5$ gray levels) {\it Prior}, Hamming {\it Likelihood} and 
Wolff cluster algorithm
for estimating $\Theta_{{\rm MAP}}$, for a ROBOT image. 
(Left) $\widehat{\Theta}$ (middle) $X$ (Right) MAP estimate.
image processing has been carried out at temperature $T=0.5$}}
\end{figure}

\section*{Discussions }
\rhead[Markov Chain Monte Carlo Image Restoration]{\thepage}
We have presented a brief review of  Markov chain Monte Carlo methods for restoration of 
digital images. There are three stages. The first consists of modeling of the 
degradation process by a conditional distribution called the {\it Likelihood}.
The data on the given corrupt image is incorporated in the {\it Likelihood}.
The second stage consists of constructing $\grave{{\rm a}}$ priori distribution.
Our expectations of how a true image should look like are modeled in the {\it Prior}.
Bayes theorem combines the {\it likelihood} and the {\it Prior} into a {\it Posterior}.
The third stage consists of sampling images from the {\it Posterior} distribution,
and making a statistical estimate of the true image. This is where Monte Carlo
methods come in. 

We have described {\it Likelihoods} based on Poisson model, Kullback-Leibler entropy distance,
Hamming distance, binary symmetric channel and Gaussian channel. The {\it Prior} model
discussed include Ising and  Potts spin. We have given a simple explanation of how 
does a {\it Posterior} maximization lead to image restoration and described a 
Monte Carlo algorithm that does this.

Metropolis algorithm  and Gibbs sampler are elegant techniques for sampling images from 
a {\it Posterior} distribution. The strategy is to start with a guess image and 
generate a Markov chain of images. We take a time homogeneous  Markov transition matrix that obeys 
detailed balance with respect to the desired {\it Posterior distribution}. Detailed balance
ensures asymptotic convergence of the Markov chain to the {\it Posterior} ensemble. Both 
Metropolis and Gibbs sampler obey detailed balance condition. The topics of 
general Markov chain, time homogeneous Markov chain, Markov transition matrix,
balance equations, balance and detailed balance conditions, asymptotic convergence, 
time reversal of Markov chain,
{\it etc} were 
discussed in the Appendix 2. In the main text, the steps involved in the implementation of 
Markov chain Monte Carlo algorithm were presented. 
\rhead[Discussions]{\thepage}
\rhead[Markov Chain Monte Carlo Image Restoration]{\thepage}

Then we took up the issue of cluster 
algorithms often employed in the context of second  order phase transition. Cluster algorithms
give you images that belong to  the {\it Prior} ensemble. 
To generate a {\it Posterior} ensemble, we sample the gray level of all the pixels in a cluster
randomly and independently from the {\it Likelihood} distribution. 
We have described  Swendsen-Wang and Wolff cluster algorithms. 

We can say that the techniques of simulating a canonical ensemble in statistical mechanics
including cluster algorithms  
have  been adapted to stochastic image restoration, 
with a reasonably good measure of success.
Recently, in statistical mechanics, non-Boltzmann ensembles 
like multicanonical or entropic ensembles 
\cite{BABTN_1991,BABTN_1992,JL},
have become popular.  An advantage of such non-Boltzmann sampling techniques
is that from  a single 
simulation we can get information over  a wide range of temperature 
by suitable reweighting schemes.  
The microstates that occur rarely in a 
canonical ensemble would occur as often as any other microstates 
in a multi canonical
or entropic ensemble. The important point is that the multicanonical
and entropic sampling Monte Carlo were
intended  for overcoming the problem of super-critical slowing 
down in first order transition. In the context of image restoration the transition 
from {\it Prior} dominated phase to {\it Likelihood} dominated phase 
is first order \cite{JMPADB}. Hence we contend that  
multicanonical/entropic sampling would have a  potential 
application in image restoration.  Work in this direction is in progress
and would be reported soon.

Investigating  algorithms based on 
invasion percolation \cite{JM1,JM2,GF} and probability changing clusters
\cite{YT1,YT2} may also prove useful in  image restoration.
The reference noisy image $X$ acts as an 
inhomogeneous external field. 
This is analogous to an Ising magnetic system in which the transition 
becomes first order due to  presence of external field.   
It would be interesting to develop cluster 
algorithms that incorporate the presence of external fields
in statistical mechanical models and adapt them to 
image processing. 
Also in stochastic image processing, ideas from random field Ising 
models may prove useful.

We can add to the list further. But then we stop here.
We conclude by saying  that analogy between image processing 
and statistical mechanics is rich and exploring  common elements 
of these two disciplines should help toward developing 
efficient image processing algorithms.  

\section*{Acknowledgments}
We  thank Baldev Raj for his keen interest, 
encouragement and enthusiastic support. One of the authors (KPN)
is thankful to 
V. S. S. Sastry for discussions on all the issues discussed in this review.
Thanks to V. Sridhar for carrying out independent simulations for testing 
preliminary versions of the algorithms reported in this review. 
KPN is thankful to S. Kanmani for discussions on the metric properties of 
distance measures and  to M. C. Valsakumar 
for discussions on Kullback-Leibler divergence
and cluster partitions.
\newpage
\section*{Appendix 1}
\rhead[Appendix]{\thepage}
\rhead[Markov Chain Monte Carlo Image Restoration]{\thepage}
\centerline{\Large{{\bf A Quick Look at Bayes' Theorem}}}
\vskip 3mm
\begin{figure}[hp]
\centerline{\includegraphics[height=2.5in,width=2.5in]{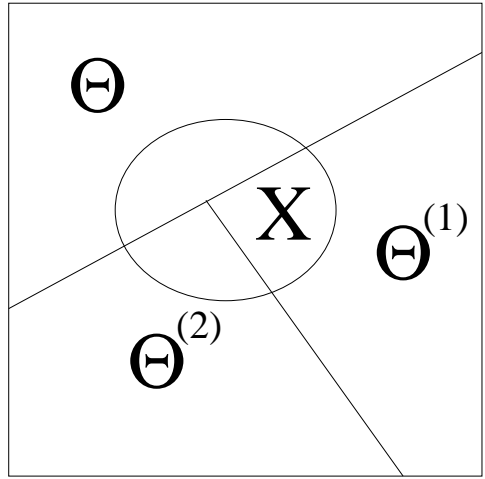}}
\caption{\small{\bf Illustration of Bayes' theorem}}
\end{figure}
\vskip 5mm
\noindent
Consider  mutually exclusive and exhaustive events
$\Theta,\ \Theta^{(1)},\ \Theta^{(2)}$.\\
We have,
\begin{eqnarray}
\Theta\cap \Theta^{(1)}  = \Theta\cap \Theta^{(2)}=\Theta^{(1)}\cap\Theta^{(2)}=\phi
\nonumber
\end{eqnarray}
and
\begin{eqnarray}
\Theta\cup\Theta^{(1)}\cup\Theta^{(2)}  \ =  \Omega.\nonumber
\end{eqnarray}
Let
$X\subset \Omega$. We have,
\begin{eqnarray}
 {\rm {\it Prior}}&:& \pi(\Theta)\ \ \ \ \ \ \ \ \ \ \nonumber\\
{\rm {\it Posterior}}&:& \pi(\Theta\vert X)\nonumber\\
{\rm  {\it Likelihood}}&:&
{\cal L}(X\vert \Theta)\nonumber
\end{eqnarray}

\begin{eqnarray}
{\cal L}(X\vert \Theta)&=&\frac{{\cal P}(X\bigcap \Theta)}{\pi(\Theta)}
\nonumber\\
\ & & \nonumber\\
\ & & \nonumber\\
\pi(\Theta\vert X)&=&\frac{{\cal P}(X\bigcap \Theta)}{P(X)}\nonumber\\
\ & & \nonumber\\
     & = &      \frac{ {\cal L} (X\vert \Theta)\pi(\Theta)}{P(X)}
\nonumber
\end{eqnarray}
$$P(X)={\cal L}(X\vert\Theta)\pi(\Theta)+{\cal L}(X\vert\Theta^{(1)})\pi(\Theta^{(1)})+
{\cal L}(X\vert\Theta^{(2)})\pi(\Theta^{(2)})$$
\newpage
\section*{Appendix 2}
\centerline{\Large{{\bf A Quick Look at Markov Chains}}}

\vskip 3mm
\subsection*{Joint and Conditional Distributions}
Consider a system which can be in any of the states denoted by
\begin{eqnarray}
{\cal A} = \{ a_1,\ a_2\ \cdots\}.
\nonumber
\end{eqnarray}
Starting from an initial state  $b_0\in{\cal A}$ at time $0$, the system
visits in successive time steps the states,
\begin{eqnarray}
b_1\in {\cal A},\  b_2\in {\cal A},\  \cdots ,\ b_n\in{\cal A},\ \cdots.\nonumber
\end{eqnarray}
The sequence of states $b_0 \in {\cal A},\ b_1\in {\cal A},\ \cdots$ are
random.
Let $P(b_n,\ b_{n-1},\ \cdots\ b_1,\ b_0)$ be the joint probability.
From the definition of conditional probability,
\begin{eqnarray}
P( b_n,\ b_{n-1},\  \cdots ,\ \ b_0) &=&
P(b_n\vert b_{n-1},\ \cdots\ ,\  b_0 )\nonumber\\
& \times & P(b_{n-1},\cdots\ ,\  b_0)\nonumber
\end{eqnarray}
Iterating we get,
\begin{eqnarray}
P(b_n,\  b_{n-1},\  \cdots ,\  b_0) & = &
                      P(b_n\vert\  b_{n-1},\  \cdots ,\   b_0)\nonumber\\
& \times& P(b_{n-1}\vert\  b_{n-2},\  \cdots ,\   b_0)\nonumber\\
& \times & P(b_{n-2}\vert\  b_{n-3},\  \cdots ,\   b_0)\nonumber\\
& \times & \cdots \nonumber\\
&\times  & P(b_2\vert b_1,\ b_0)\nonumber\\
&\times & P(b_1\vert b_0)\nonumber\\
&\times&P(b_0)\nonumber
\end{eqnarray}
\noindent
\subsection*{Markovian Assumption}
The sequence of states
$b_0,\ b_1,\ \cdots\ ,\ b_n$ constitutes a Markov chain, if
\begin{eqnarray}
P(b_k\vert b_{k-1},\ b_{k-2},\ \cdots,\ b_1,\ b_0)=P(b_k\vert b_{k-1})
\ \forall\ k=1,\ n
\nonumber
\end{eqnarray}
As a consequence we have,
\begin{eqnarray}
P(b_n,\ b_{n-1},\  \cdots ,\ b_1,\
 b_0) = P(b_0)\prod_{k=1}^{n}P(b_k\vert b_{k-1})\nonumber
\end{eqnarray}
The state of the system at time step $k+1$ depends only on its present state
(at time step $k$) and not on the states it visited at all the previous time steps,
$k-1,$  $k-2,$ $\cdots,$ $2,$  $1,$ $0$.
{\it Future is independent of the past once the present is specified.}
We call $P(b_{k+1}\vert b_k)$ the transition probability at time $k$.
In general the transition probability depends on $k$.
\subsection*{Stationary Markov Chain}
We specialize to the case when the transition probability is independent
of the time index $k$. Then we  get a  homogeneous or stationary Markov chain.
We have therefore,
\begin{eqnarray}
P(b_{k+1} = a_i\vert b_k = a_j) = M_{i,j}\ \forall\ k\nonumber
\end{eqnarray}
Thus a (homogeneous) Markov chain is completely defined once we specify,
\begin{enumerate}
\item[-]the state space ${\cal A} = \{ a_1,\ a_2,\ \cdots\}$
\item[-] the  probabilities $\{ M_{i,j}\ :\  \ \forall\  i,\ j\}$ for transition 
from one state ($a_j\in {\cal A}$)  to another state ($a_i\in {\cal A}$),
 and
\item[-]the initial state $b_0\in{\cal A}$
\end{enumerate}
\subsection*{Markov Transition Matrix}
$M$, defined above,  is called the transition matrix.
For every  state $a_j$, because of normalization, we have,
$$\ \sum_i P(a_i\vert a_j) = \sum_i M_{i,j} = 1$$
In other words, the elements of each column of
the transition matrix $M$ add to unity
A matrix with non-negative elements and for which
the elements of each column add to unity is called a  Markov matrix

Let $\langle U\vert$ denote the uniform unnormalized probability
vector  $(1,\ 1,\ \cdots\ ,\ 1)$.
It is easily seen that
$\langle U\vert$ is the left eigenvector of  $M$ corresponding to
the eigenvalue unity: 
$$\langle U\vert M=\langle U\vert .$$
The right eigenvector of $M$ corresponding to
eigenvalue unity is called the invariant distribution 
of $M$, and is
denoted by $\vert\pi\rangle$: $M\vert \pi\rangle=\vert \pi\rangle$;
the largest eigenvalue of $M$ is real , unity and non-degenerate 
$\Rightarrow$ $\vert\pi\rangle$ is unique. 

The eigenvectors of $M$ form a complete set.
Let $\vert\pi_0\rangle$ be an arbitrary initial vector with non-zero overlap
with $\vert\pi\rangle$ {\it i.e.} $\langle\pi\vert\pi_0\rangle\ne 0$.
We express
\begin{eqnarray}
\vert\pi_0\rangle = \vert\pi\rangle\langle\pi\vert\pi_0\rangle+
\sum_{\lambda\ne 1}\vert\lambda\rangle\langle\lambda\vert\pi_0\rangle\nonumber
\end{eqnarray}
where $\vert\lambda\rangle$ is the eigenvector corresponding to eigenvalue
$\lambda\ne 1$ and $\vert\lambda=1\rangle=\vert\pi\rangle$
\begin{eqnarray}
M^n\vert\pi_0\rangle &=& \vert\pi\rangle\langle\pi\vert\pi_0\rangle+\sum_{\lambda\ne 1}\lambda^n\vert\lambda\rangle
\langle\lambda\vert\pi_0\rangle\nonumber\\
&{}^{\ \ \sim}_{n\to\infty}&\vert\pi\rangle\ \ \ {\rm since}\ \ \ \vert\lambda\vert^n {}^{\ \ \ \sim}_{n\to\infty}\ \ 0\nonumber
\end{eqnarray}
\vskip 3mm

\noindent
How do we construct a Markov matrix $M$ whose invariant vector is the
desired distribution $\vert \pi\rangle$ ?

\subsection*{Balance Condition}
\noindent
Let $P(a_i,n)$ denote the probability that the system is in state
$a_i$ at discrete time $n$. In other words $P(a_i,n)$ is the probability that
$b_n=a_i$.
Formally we have the discrete time {\bf balance}  equation
\begin{eqnarray}
P(a_i,n+1)&\negthickspace\negthickspace = \negthickspace\negthickspace &
\sum_{j\ne i}M_{i,j}P(a_j,n)+\bigg( 1-\sum_{j\ne i} M_{j,i}\bigg) P(a_i,n)\nonumber\\
\ \nonumber\\
& = &
\sum_j\bigg[ M_{i,j}P(a_j,n)-M_{j,i}P(a_i,n)\bigg] + P(a_i,n)\nonumber
\end{eqnarray}
We need $P(a_i,n+1)=P(a_i,n)=\pi_i\ \forall\  i$ when
$n\to\infty$ for asymptotic equilibrium.
We can ensure this by demanding that the sum over $j$ of the terms in the
right hand side  of the master equation be zero with respect to 
\begin{eqnarray}
 \pi_i = {}^{\ {\rm Lim.}}_{n\to\infty}\ P(a_i,n)\ \ \forall\ \ i\nonumber
\end{eqnarray}
In other words, we demand 
\begin{eqnarray}
\sum_j \Big[ M_{i,j}\pi_j-M_{j,i}\pi_i\Big]=0.\nonumber
\end{eqnarray}
This is called a balance condition.
The balance condition can be written as 
\begin{eqnarray}
M\vert\pi\rangle =\vert\pi\rangle .
\nonumber
\end{eqnarray}

\subsection*{Detailed Balance Condition}
It is difficult to construct an $M$ whose asymptotic distribution is the desired $\vert\pi\rangle$ employing
the balance condition.
Hence we employ a more restrictive {\bf detailed balance} condition by demanding each term in the sum be zero:
\begin{eqnarray}
\pi_j M_{i,j}=\pi_i M_{j,i}.\nonumber
\end{eqnarray}

The physical significance of the detailed balance is that the corresponding equilibrium Markov chain
is time reversible. This means that it is impossible to tell whether a movie of  a sample path
of images is being shown forward or backward.
 
\subsection*{Time Reversal or $\pi$ Dual of $M$}
Consider a time homogeneous Markove chain denoted by
$$b_o\to b_1\to \cdots \to b_{k-1} \to b_{k}  \to b_{k+1}\to \cdots b_N.$$
Let us consider a chain whose entries are in the reverse order and denote it 
$$\widehat{b}_o\to \widehat{b}_1\to \cdots \to \widehat{b}_{k-1} \to 
\widehat{b}_{k}  \to \widehat{b}_{k+1}\to \cdots \widehat{b}_N,$$
where $$\widehat{b}_k = b_{N-k}.$$
Notice that the forward and the reversed chain have entries belonging to the 
same state space ${\cal A}$. The reverse chain is also stationary and
 Markovian 
and hence is completely specified by the state space ${\cal A}$, 
a Markov transition matrix denoted by $\widehat{M}$ and 
an initial state $\widehat{b}_0=b_N$. Aim is to express $\widehat{M}$ in terms 
of $M$ and $\vert \pi\rangle$.  
 
To this end, we define a joint probability matrix $W$ and express it in terms
of $M$ and $\vert \pi\rangle$. We define
$$ W_{i,j}=P(b_{n+1}=a_i,b_{n}=a_j)\ \forall\ \ \  i,j$$ 
Since we are considering stationary Markov chain $W$ is independent of
the time index $n$, except for the time ordering
.
 In general one must consider a time reversal of transition matrix. Since in an equilibrium
system a forward transition and its reverse occur with the same probability, we can write formally,
\begin{eqnarray}
\pi_i M_{j,i}=\pi_j \widehat{M}_{i,j}\nonumber
\end{eqnarray}
where $\widehat{M}$ is the time reversal or the $\pi$-dual of $M$, given by,
\begin{eqnarray}
\widehat{M}={\rm diag}\ (\{\pi_i\})M^{\dagger}{\rm diag}\ (\{1/\pi_i\})\nonumber
\end{eqnarray}
where ${\rm diag}(\{ \eta_i\})$ denotes a diagonal matrix whose (diagonal) elements 
are $\{ \eta_i\}$ and $M^{\dagger}$ is the transpose of $M$. The matrix $\widehat{M}$ is  Markovian;
its invariant vector is the same as that of $M$. If a transition matrix obeys detailed balance
then $M=\widehat{M}$.
\vskip 3mm
\noindent
\subsection*{Metropolis Algorithm}
$$M_{i,j} = \alpha\ {\rm min} \Big( 1,\frac{\pi_i}{\pi_j}\Big)$$
where $\alpha$ is a convenient
constant introduced for ensuring normalization:\\
$\sum_i M_{i,j} =1\ \forall\ \ j.$

\vskip 5mm
\noindent
\subsection*{Glauber/Heat-Bath/Gibbs Sampler}
\rhead[References]{\thepage}
\rhead[Markov Chain Monte Carlo Image Restoration]{\thepage}
$$M_{i,j}=\alpha\ \Big( \frac{\pi_i}{\pi_i +\pi_j}\Big)$$
\vskip 5mm
\noindent
It is easily verified that the Metropolis algorithm and the Glauber/Heat-bath/Gibbs algorithm obey
detailed balance condition;
also only the ratio of the probabilities appear in these algorithms;
It is this property that comes  handy
in generation of a Markov chain for a statistical mechanical system wherein the 
normalization called the partition function is not known. Hence for the 
Metropolis and Gibbs sampler, it is adequate if we know the desired 
target distribution up to a normalization constant. 
 

\vskip 5mm
\begin{thebibliography}{1000}
\bibitem{DECG}
H. Derin, H. Elliot, R. Cristi and D. Gemen,
{\it Bayes smoothing algorithms for segmentation
of binary images modeled by Markov random fields}, 
IEEE Transactions of Pattern Analysis and Machine Intelligence,
PAMI-{\bf 6} 707 (1984)
\bibitem{SGDG}
S. Gemen and D. Gemen,
{\it Stochastic Relaxation, Gibbs distributions and Bayesian Restoration},
IEEE Transactions of Pattern Analysis and Machine Intelligence,
{\bf 6} 721 (1984)
\bibitem{JEB_1986}
J. E. Besag,
{\it On the statistical analysis of dirty pictures (with discussions)},
J. Royal Statistical Society,
B {\bf 48} 259 (1986)
\bibitem{GW}
G. Winkler,
{\it Image analysis, random fields and dynamic Monte 
Carlo methods},
Springer, Berlin (1995)
\bibitem{KT}
K. Tanaka,
{\it Statistical Mechanical Approach to Image Processing},
J. Phys. A:Math. Gen., {\bf 35} R81 (2002)
\bibitem{JMPADB}
J. M. Pryce and A. D. Bruce,
{\it Statistical mechanics of image restoration},
J. Phys. A: Math. Gen. A {\bf 28}, 511 (1995)
\bibitem{HN}
H. Nishimori, {\it Statistical physics of spin glasses and information 
processing: an introduction}, Oxford university Press, Oxford (2001)
\bibitem{HNKMYW}
H. Nishimori and K. M. Y. Wong, {\it Statistical mechanics of image restoration and error-correcting codes},
Phys. Rev. E {\bf 60}, 132 (1999).
\bibitem{KPN_Madurai}
K. P. N. Murthy, {\it Bayesian restoration of digital images employing 
Ising and Potts Priors}, invited talk at the National Seminar on Recent
Trends in Digital Image Processing and Applications, Yadava College,
Madurai 28 - 29, October 2004. See Proceedings (2004)p.1
\bibitem{Kullback}
S. Kullback, {\it Information Theory and Statistics},
Wiley New York (1968)
\bibitem{RLD}
R. L. Dobrushin, {\it The description of a random field
by means of conditional probabilities, and conditions of its 
regularity}, Theory Prob. Appl., {\bf 13}, 197 (1968) 
\bibitem{JEB_1974}
J. E. Besag, {\it Spatial interactions and the statistical analysis of 
lattice systems (with discussions)},
J. Royal Statistical  Society, {\bf 36} 192 (1974)
\bibitem{OFDS}
O. Frank and D. Strauss, {\it Markov graphs}, 
J. Amer. Stat. Assoc., {\bf 81}, 832 (1986)
\bibitem{JMHPC}
J. M. Hammersley and P. Clifford, {\it Markov fields of finite graphs and lattices}, University of California, Berkeley (1968): cited in \cite{JMPADB}
\bibitem{Ising}
E. Ising, 
Zeitschrift Physik,
{\bf 31}, 252 (1925);
S. G. Brush, {\it History of the Lenz-Ising model}, Rev. Mod. Phys., {\bf 39}, 883 (1967)
\bibitem{Potts}
R. B. Potts,
Proc. Cambridge Phil. Soc., {\bf 48}, 106 (1952)
\bibitem{SGDEM_1987}
S. Gemen and D. E. McClure,
{\it Statistical methods in tomographic image reconstruction},
Proc. 46-th session of the International statistical institute,
Bulletin of ISI, {\bf 52}, 5 (1987)
\bibitem{Feller}
W. Feller, 
{\it An introduction to Probability theory and its applications},
Wiley Third edition Vol. 1 (1970)p.124
\bibitem{Papoulis}
A. Papoulis, {\it Probability, random variables and stochastic processes},
McGraw-Hill Kogakusha Ltd. (1965)p.38
\bibitem{Metropolis}
N. Metropolis, A. W. Rosenbluth, M. N. Rosenbluth, A. H. Teller and E. Teller,
{\it Equation of state calculation by fast computing machine},
J. Chem. Phys., {\bf 21}, 1087 (1953)
\bibitem{NMSU}
N. Metropolis and S. Ulam, {\it The Monte Carlo Method},
J. Amer. Statistical Assoc., {\bf 44}, 335 (1949)
\bibitem{JMHDCH}
J. M. Hammersley and D. C. Handscomb, {\it Monte Carlo Methods}, Chapman and
Hall, London (1964)
\bibitem{DPLKB}
D. P. Landau and K. Binder,
{\it A Guide to Monte Carlo Simulations in Statistical Physics},
Cambridge University Press (2000)
\bibitem{IMS}
I. M. Sobol, {\it The Monte Carlo Method}, Mir, Moscow (1975)
\bibitem{FJ}
F. James, {\it Monte Carlo theory and practice}, Rep. Prog. Phys., {\bf 43}
1145 (1980)
\bibitem{KBDWH}
K. Binder and D. W. Heermann, {\it Monte Carlo Simulation in Statistical
Physics: An Introduction}, Springer (1988)
\bibitem{PC}
Paul Coddington, {\it Monte Carlo simulation for Statistical
Physics}, Report CPS-713, Northeast Parallel Architectures (1996)
\bibitem{KPNM_2000}
K. P. N. Murthy, {\it Monte Carlo: Basics}, Monograph ISRP/TD-3,
Indian Society  for Radiation Physics (2000). (eprint: arXiv: cond-mat/0104215
v1 12  April 2001
\bibitem{KPNM_2004}
K. P. N. Murthy,
{\it Monte Carlo Methods in Statistical Physics},
Universities Press (India) Private Limited,
distributed by Orient Longmann private Limited (2004)
\bibitem{WKH}
W. K. Hastings,
{\it Monte Carlo sampling methods using Markov chains and their applications},
Biometrica, {\bf 57}, 97 (1970) 
\bibitem{Noris}
J. R. Noris,
{\it Markov Chains}, Cambridge University Press (1997)
\bibitem{MC}
M. Cruetz, {\it Confinement and critical dimension of space-time},
Phys. Rev. Lett., {\bf 43}, 553 (1979)
\bibitem{BDR}
B. D. Ripley, {\it Modeling spatial patterns (with discussion)},
J. Royal Statistical Society, B {\bf 39}, 172 (1977)
\bibitem{UG}
U. Grenader, {\it Tutorial in Pattern Theory}, Report: Divison of Applied  Mathematics,
Brown University (1983) 
\bibitem{Glauber}
R. J. Glauber,
{\it Time-dependent statistics of the Ising model},
J. Math. Phys., {\bf 4}, 294 (1963)
\bibitem{SW} 
R. H. Swendsen and J.-S. Wang, {\it Non-universal critical dynamics
in Monte Carlo simulation}, Phys. Rev. Lett., {\bf 58}, 86 (1987)
\bibitem{PWKCMF}
P. W. Kasteleyn, C. M. Fortuin, J. Phys. Soc. Japan Suppl., {\bf 26}, 11 (1969);
C. M. Fortuin and P. W. Kasteleyn, {\it On the random - cluster model: I.
Introduction and relation to other models},
Physica(Utrecht), {\bf 57}, 536 (1972).
\bibitem{ACWK}
A. Coniglio and W. Klein, {\it Clusters and Ising critical
droplets: a renormalization group approach}, J. Phys., A {\bf 13}, 2775 (1980)
\bibitem{AJG}
A. J. Gray,
{\it Simulating Posterior Gibbs distributions: a comparison of the Swendsen-Wang algorithm
and Gibbs sampler}, Statistics and computing, {\bf 4}, 189 (1994)
\bibitem{IG}
I. Gaudron, {\it Rate of convergence of Swendsen-Wang dynamics in image segmentation problems: a theoretical
and experimental study}, J. Stat. Phys., (1996)
\bibitem{DJNPJG}
D. J. Nott and P. J. Green,
{\it Bayesian variable selection and Swendsen-Wang algorithm}, J. Computational and graphical
statistics, {\bf 13}, 1 (2004)
\bibitem{SLLWN}
S. Lasota and W. Niemero, {\it A version of Swendsen-Wang algorithm for restoration of images degraded 
by Poisson noise}, Pattern Recognition, {\bf 36}, 931 (2003)
\bibitem{RGEADS}
R. G. Edwards, and A. D. Sokal, {\it Generalization of Fortuin-Kasteleyn-Swendsen-Wang representation and Monte Carlo
algorithm}, Phys. Rev.,  D {\bf 38}, 2009 (1988)
\bibitem{JBPJG}
J. Besag and P. J. Green,
{\it Spatial statistics and Bayesian computation (with discussion)}
J. Royal Statistical Society,  B {\bf 16}, 395 (1993)
\bibitem{DMH}
D. M. Hidgon, {\it Auxiliary variable methods for Markov chain Monte Carlo with applications},
J. Amer. Statistical Association, {\bf 93}, 585 (1998)
{\it Collective Monte Carlo updating for spin systems},
Phys. Rev. Lett., {\bf 62}, 361 (1989)
\bibitem{Wolff}
U. Wolff, {\it Collective Monte Carlo updating for spin systems},
Phys. Rev. Lett. {\bf 62} 361 (1989)
\bibitem{BABTN_1991}
B. A. Berg and T. Neuhaus, {\it Multicanonical
algorithms for first order phase transition},
Phys. Lett. B {\bf 267 }, 249 (1991)
\bibitem{BABTN_1992}
B. A. Berg and T. Neuhaus, {\it Multicanonical ensemble: a new approach
to simulation of first order phase transition},
Phys. Rev. Lett. {\bf 68}, 9 (1992)
\bibitem{JL}
J. Lee, {\it New Monte Carlo algorithm: entropic
sampling}, Phys. Rev. Lett. {\bf 71}, 211 (1993); Erratum: {\bf 71} 2353 (1993)
\bibitem{JM1}
J. Machta, Y. S. Choi, A. Lucke, T. Schweizer
 and L. M. Chayes, {\it Invaded cluster algorithm for equilibrium
critical points}, Phys. Rev. Lett. {\bf 75}, 2792 (1995)
\bibitem{JM2}
J. Machta, Y. S. Choi, A. Lucke, T. Schweizer and L. M. Chayes,
{\it Invaded cluster
algorithm for Potts models}, Phys. Rev. E {\bf 54}, 1332 (1996)
\bibitem{GF}
G. Franzese, V. Cataudella and A. Coniglio,
{\it Invaded cluster dynamics for frustrated models},
Phys.  Rev. E {\bf 57}, 88 (1998)
\bibitem{YT1}
Y. Tomita and Y. Okabe, {\it Probability changing cluster
algorithm for Potts model}, Phys. Rev. Lett. {\bf 86}, 572 (2001)
\bibitem{YT2}
Y. Tomita and Y.~Okabe, {\it Probability-changing cluster
algorithm: study of three - dimensional
Ising model and percolation  problem}, eprint: arXiv: cond-mat/0203454 (2002)

\end{thebibliography}
\end{document}